\documentclass[journal]{IEEEtran}
\usepackage{graphicx}
\usepackage{booktabs}
\usepackage{subfigure}
\usepackage{amsmath}
 \usepackage{amssymb}
\usepackage{multirow}
\usepackage{makecell}
\usepackage{threeparttable}
\usepackage{graphicx}
\usepackage{color, xcolor}
\usepackage{stfloats}
\usepackage{multirow}
\usepackage{hyperref}
\hypersetup{
    colorlinks=true,
    linkcolor=blue,
    urlcolor=blue,
    citecolor=blue
}

\begin{document}

\title{Aerial Multi-View Stereo via Adaptive Depth Range \\ Inference and Normal Cues}

\author{
Yimei Liu,
Yakun Ju, \emph{Member, IEEE},
Yuan Rao, 
Hao Fan, 
Junyu Dong, \emph{Member, IEEE}, 
Feng Gao, \emph{Member, IEEE},
Qian Du, \emph{Fellow, IEEE}

\thanks{This work was supported in part by the National Science and Technology Major Project of China under Grant 2022ZD0117202, and in part by the National Natural Science Foundation of China under Grant 42106193. (\textit{Corresponding authors}: Junyu Dong and Feng Gao)}
\thanks{Y. Liu, Y. Rao, H. Fan, J. Dong and F. Gao are with the Department of Information Science and Technology, Ocean University of China, Qingdao 266100, China (email: liuyimei@stu.ouc.edu.cn, raoyuan@ouc.edu.cn, fanhao@ouc.edu.cn, dongjunyu@ouc.edu.cn, gaofeng@ouc.edu.cn)

Y. Ju is with School of Computing and Mathematical Sciences, University of Leicester. (email: yj174@leicester.ac.uk)

Qian Du is with the Department of Electrical and Computer Engineering, Mississippi State University, Starkville, MS 39762 USA. (email: du@ece.msstate.edu)}}

\markboth{IEEE Transactions on Geoscience and Remote Sensing}
{Shell}

\maketitle

\begin{abstract}
Three-dimensional digital urban reconstruction from multi-view aerial images is a critical application where deep multi-view stereo (MVS) methods outperform traditional techniques. However, existing methods commonly overlook the key differences between aerial and close-range settings, such as varying depth ranges along epipolar lines and insensitive feature-matching associated with low-detailed aerial images. To address these issues, we propose an Adaptive Depth Range MVS (ADR-MVS), which integrates monocular geometric cues to improve multi-view depth estimation accuracy. The key component of ADR-MVS is the depth range predictor, which generates adaptive range maps from depth and normal estimates using cross-attention discrepancy learning. In the first stage, the range map derived from monocular cues breaks through predefined depth boundaries, improving feature-matching discriminability and mitigating convergence to local optima. In later stages, the inferred range maps are progressively narrowed, ultimately aligning with the cascaded MVS framework for precise depth regression. Moreover, a normal-guided cost aggregation operation is specially devised for aerial stereo images to improve geometric awareness within the cost volume. Finally, we introduce a normal-guided depth refinement module that surpasses existing RGB-guided techniques. Experimental results demonstrate that ADR-MVS achieves state-of-the-art performance on the WHU, LuoJia-MVS, and München datasets, while exhibits superior computational complexity. 
\end{abstract}


\section{Introduction}

\IEEEPARstart{L}arge-scale 3D reconstruction of the Earth's surface has traditionally relied on commercial software\footnote{http://www.bentley.com/en/products/brands/contextcapture,}\footnote{http://www.pix4d.com/} using conventional dense matching methods~\cite{patchmatch} \cite
{hir2007stereo} \cite{mahato2019dense}. As more recent works have demonstrated that Multi-View Stereo (MVS) can be better solved with deep learning techniques for close-range object reconstruction~\cite{2018MVSNet} \cite{RMVSNet} \cite{2020PatchmatchNet} \cite{giang2021curvature} \cite{Aanaes2016large} \cite{Yao_BlendedMVS_2020} \cite{Kna2017tanks}, learning-based MVS tailored for large-scale aerial images have gained significant attention in the research community~\cite{li2023hierarchical} \cite{REDNet} \cite{egmvs2024} \cite{rao2020bidirectional}. A key challenge in aerial vs. close-range MVS is the difference in depth ranges projected onto epipolar lines, making multi-view matching more intractable. Methods designed for close-range images often cannot transfer well to aerial scenarios, despite being trained on aerial stereo datasets.

\begin{figure*}[!h]
\begin{center}
\includegraphics[width=0.7\linewidth]{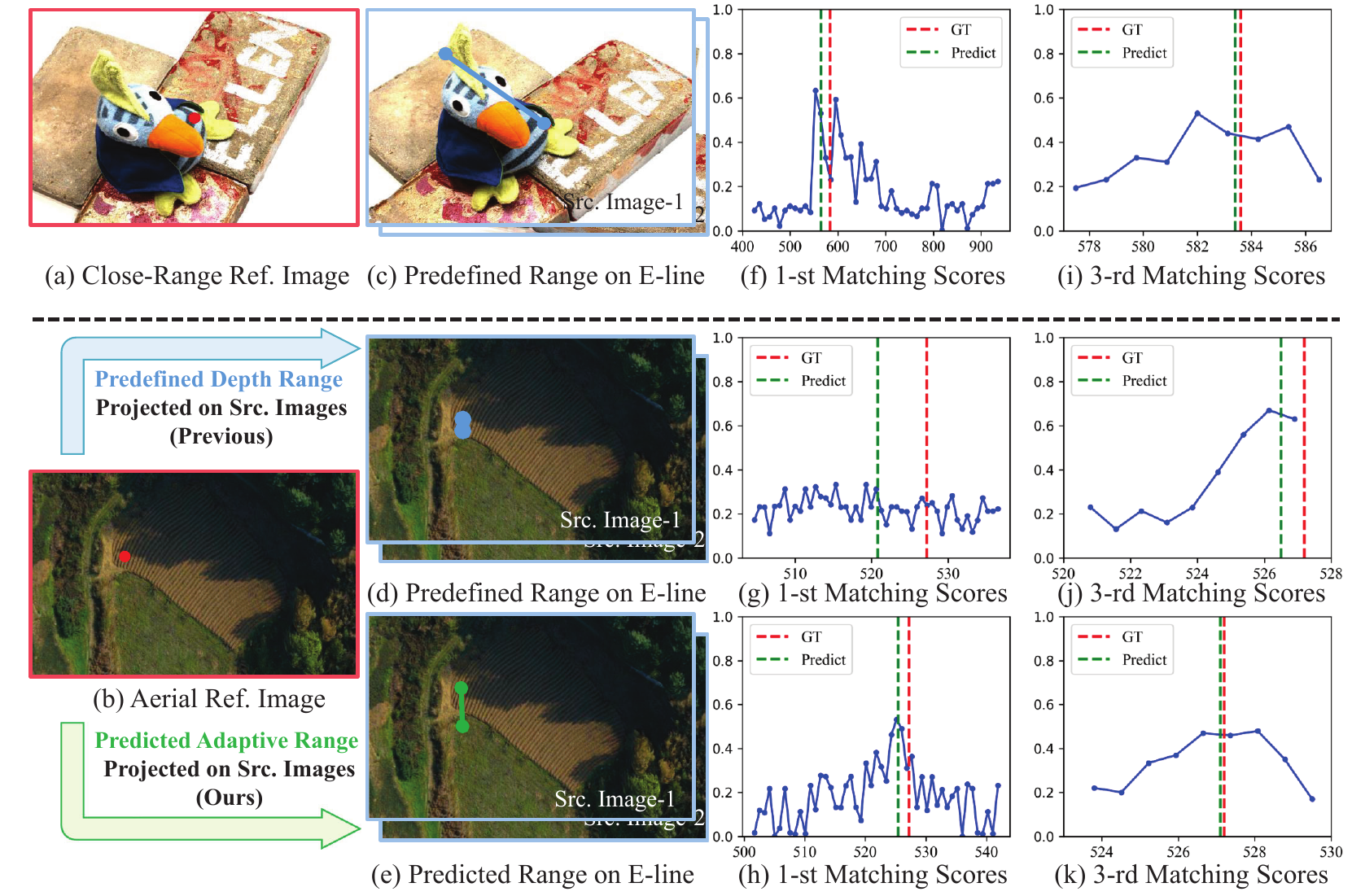}
\end{center}
\caption{Feature matching scores for close-range vs. aerial image pairs. For a given pixel in the reference image (a-b), the initial depth ranges projected onto the Epipolar line (E-line) of the source view are shown in (c-e). Feature matching scores across different depth samples are presented for the first stage (f-h) and the last stage (i-k). The x-axis represents depth sample values uniformly distributed within the depth range (in meters), and the y-axis indicates feature matching scores. For the close-range scenario, scores are derived from the DTU-trained GoMVS model~\cite{wu2024gomvs} \cite{Aanaes2016large}, while for the aerial scenario, scores are produced by the Luojia-MVS-trained GoMVS model and our Luojia-MVS-trained model~\cite{li2023hierarchical}.}
\label{fig:teaser} 
\end{figure*}

Current learning-based MVS approaches typically employ cascaded cost volumes for depth estimation. In this framework, a predefined depth range, derived from projecting sparse 3D points generated by preliminary Structure-from-Motion (SfM), is first used for uniform depth candidate sampling, followed by feature matching, cost aggregation, and depth regression. Then it progressively refines depth estimation by narrowing the sampling range and increasing the spatial resolution of cost volumes, ultimately producing multi-resolution depth maps from coarse to fine scales.
However, as illustrated in Fig.~\ref{fig:teaser}, while predefined depth range sampling is effective for close-range image pairs, it becomes less reliable for aerial imagery. This limitation stems from the significantly greater capture distance in aerial images, which reduce displacement along the epipolar line and lead to less discriminative feature matching across depth samples. Consequently, the estimated depth may deviate far from the actual depth and cannot be corrected by local depth searches in later stages, resulting in inaccurate predictions. Therefore, we argue that an adaptively expanded depth range is essential for constructing a more discriminative initial cost volume and is crucial for improving depth estimation accuracy in aerial MVS.

The recent surge in monocular geometric estimation methods, driven by significant scale-up in data availability, presents an opportunity to complement feature matching-based stereo methods. Motivated by this, we propose leveraging monocular geometric cues to learn adaptively enlarged depth range maps, which serve as an initial prior to guide matching-based depth estimation. 
However, as evidenced by statistics on the aerial dataset~\cite{li2023hierarchical} (Table~\ref{tab:mono_sta}), due to the ill-posed nature, the monocular estimation alone, after directly scaled to predefined ranges, falls short of achieving the meter- or centimeter-level precision required for high-quality 3D reconstruction. More importantly, we notice that monocular depth estimates often exhibit spurious variations caused by uneven brightness or texture interference. In contrast, monocular normal maps are less susceptible to such issues, but reveal only the relative depth variation.

Based on these observations, we propose integrating complementary monocular depth and normal cues to enhance aerial MVS accuracy. First, we propose a depth range predictor to improve cascaded cost formulation. The rationale behind our design can be summarized in three key points: (1) Depth and normal maps are inherently correlated, both offering strong geometric cues to guide cost volume formulation in MVS; (2) We hope to sample wider depth ranges in regions with sudden depth changes and larger normal variations, which is beneficial for accurate depth estimation in these edge regions; (3) Since normal maps reflect relative depth variation rather than absolute precision, they do not require high precision values for depth range prediction. Thus, we allow monocular normal estimates to remain unchanged across stages. Specifically, we first directly scaled monocular depth as an initial estimate and predict depth range maps through cross-attention feature interaction within monocular geometric cues. In subsequent stages, the monocular depth is replaced by regressed depth maps, while the monocular normal remains unchanged. Consequently, per-pixel adaptive depth range maps are predicted for cascaded cost volume construction, with minimized depth regression deviations at each stage. 

The experimental results show that, initially, our adaptive depth range maps typically extend beyond predefined boundaries. This design improves discriminative feature matching in aerial imagery by compensating for reduced displacement caused by greater capture distances, ensuring more accurate depth estimation. In later stages, the depth range maps are progressively refined, ultimately converging with those in standard cascaded architectures for precise depth regression.

\begin{table}[t!]
\centering
\caption{Performance comparison of monocular estimation models on LuoJia-MVS dataset~\cite{li2023hierarchical}.}
\label{tab:mono_sta} 
\resizebox{0.95\linewidth}{!}{
\begin{tabular}{c|cccc}
\hline\toprule
\textbf{Mono Prior Model} & \textbf{MAE (m)} & \textbf{MED (m)} & \textbf{MAE (°)} & \textbf{MED(°)} \\
\midrule
DepthAnythingV2~\cite{yang2024depth}    & 7.7    & 7.9    & -     & -    \\
DepthPro~\cite{DepthPro2024}& 7.4    & 7.8    & -     & -    \\
Lotus~\cite{he2024lotus}   & 7.6    & 8.1    & 18.7  & 17.5 \\
Omnidata~\cite{omnidata}    & -& -& 14.2  & 15.3 \\ 
\bottomrule\hline
\end{tabular}}
\begin{tablenotes}
\footnotesize
\item{MAE and MED denote the mean and median deviations of monocular depth and normal estimations, respectively. The GT normal are compute from the GT depth maps in closed form~\cite{qi2018geonet}.}
\end{tablenotes}
\end{table}
Furthermore, for the critical cost aggregation step, we adapted geometrically consistent cost aggregation (GCA)~\cite{wu2024gomvs}, originally developed for close-range MVS, for aerial stereo images. Like GCA, we first perform 3D depth space alignment and then aggregate relevant matching costs. Differently, we extend the fixed 3 × 3 window of GCA by applying deformable learning to the monocular normal map, alleviating insensitive matching in low-detailed aerial images. Meanwhile, considering that expanded neighboring positions may belong to disconnected planes with similar surface normals, we further incorporate feature similarity weights to ensure high-quality cost aggregation. Finally, we deigne a Normal-Guided Depth Refinement (NDR) module to upscales and refines the matching-based depth map to the input resolution, leveraging monocular normal instead of RGB-guided refinement~\cite{2018MVSNet} \cite{2020PatchmatchNet}. The monocular normal provides more stable and direct cues to enhance depth estimation accuracy in aerial MVS.

In summary, the contributions of this paper are as follows:
\begin{itemize}
\item \textbf{Depth Range Predictor (DRP):} By analyzing the gap between close-range and aerial MVS, we introduce the DRP to promote discriminative feature matching in multi-view remote sensing images. This is the first attempt to leverage large-scale monocular geometric models for aerial MVS, enabling high-precision depth estimation.
\item \textbf{Normal-Guided Cost Aggregation (NCA):} The NCA operator is designed to effectively tackle the challenge of feature matching in low-detailed aerial images, thereby improving depth estimation accuracy.
\item \textbf{Normal-Guided Depth Refinement (NDR):}  We propose the NDR module, which exploits stable and direct normal cues to further refine depth estimation.
\end{itemize}

\section{Related Work}
\subsection{Large-scale Aerial MVS}
Large-scale aerial MVS datasets, such as WHU~\cite{REDNet} and LuoJia-MVS~\cite{li2023hierarchical}, has significantly advanced the development of learning-based aerial MVS methods~\cite{gao2023general} \cite{REDNet} \cite{li2023hierarchical} \cite{mao2024sdl} \cite{egmvs2024} \cite{huang2024multi} \cite{liu2023deep} \cite{chen2024surface}. 
Many of these methods extend close-range MVS approaches to better address the distinctive features of aerial imagery. RED-Net~\cite{REDNet} was the first tailored aerial MVS method, employing stacked GRU blocks within a recurrent encoder-decoder structure for efficient cost aggregation. It extended R-MVSNet~\cite{RMVSNet} by replacing the fixed 3 × 3 regularization kernel with multi-scale receptive fields, achieving higher depth estimation accuracy than earlier traditional methods. Subsequent methods, such as HDC-MVSNet~\cite{li2023hierarchical} and SDL-MVS~\cite{mao2024sdl}, introduced cascaded MVS networks with progressively narrowed depth ranges and center-focused discretization strategy, improving depth estimation in small object regions. More recently, EG-MVSNet~\cite{egmvs2024} introduced a multi-branch two-stage architecture to jointly estimate depth and edge maps, using edge-guided refinement to enhance accuracy at building boundaries. CSC-MVS~\cite{huang2024multi} highlighted that aerial depth estimation cannot rely solely on feature matching, proposing an uncertainty-based multi-task optimization method to adaptively combine matching and semantic metrics.

While these advancements \cite{mao2024sdl} \cite{egmvs2024} have also incorporated adaptive depth ranges for cascaded cost volume construction and improved depth estimation accuracy in detailed regions, the adopted multi-branch and multi-task architectures have significantly increased computational complexity and memory usage. More critically, they overlook the fundamental difference between close-range and aerial MVS tasks, with the initial sampling range remaining predefined by sparse 3D points from SfM. 
We argue that, while this common practice may suffice for close-range images, it is inadequate for aerial imagery, where reduced displacement leads to non-discriminative feature matching and complicates depth optimization. To address this, our method predicts adaptive range maps from the depth and normal domains, expanding the initial depth range and progressively refining it in later stages to enhance depth optimization at each phase. Additionally, we integrate monocular priors at multiple steps of the matching-based depth estimation pipeline, improving cost aggregation and overall depth estimation performance without additional memory overhead.
\begin{figure*}[!tp]
\begin{center}
\includegraphics[width=0.8\linewidth]{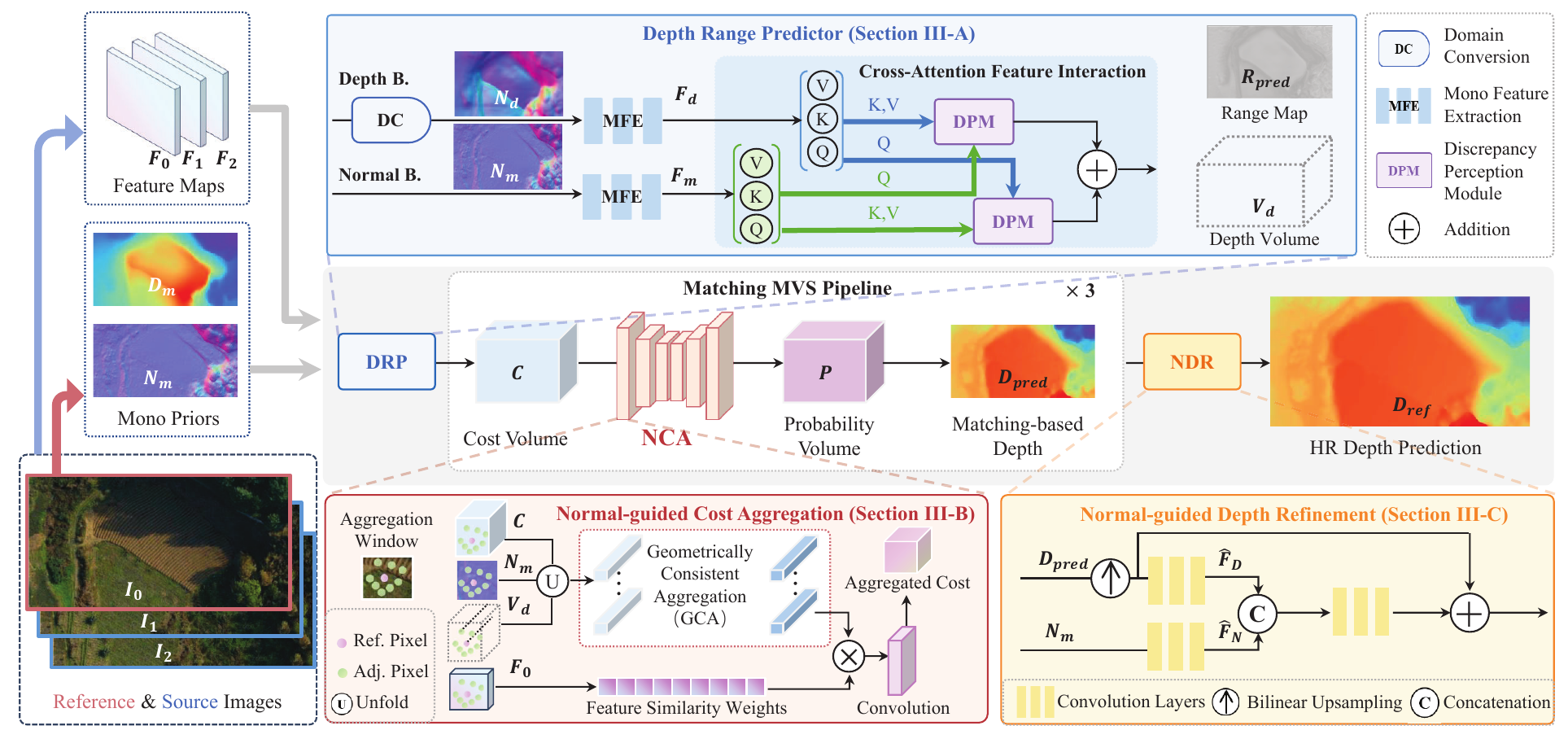}
\end{center}
\vspace*{-3mm}
\caption{The overall framework of our proposed ADR-MVS. The network consists of three primary components: depth range map predictor, matching-based depth estimation, and normal-guided depth upsampling and refinement. The feature extraction for the reference and source images, along with the GCA operation of cost aggregation network, is based on the GoMVS~\cite{wu2024gomvs}. For brevity, the feature extraction process is not elaborated in the methodology.}
\label{fig:arch} 
\end{figure*}

\subsection{Incorporating Priors into MVS}

Several studies have explored the integration of priors such as semantic metrics \cite{huang2024multi} \cite{sdmvs2024}, monocular depth~ \cite{magnet2022} \cite{bartolomei2024stereo} \cite{chen2021revealing} \cite{watson2020learning}, and normal estimates \cite{Li_NR_MVSNet} \cite{wu2024gomvs} \cite{virtualnor2019} \cite{normalstereo2020} \cite{normalocc2020} \cite{nvpamvs2022} in close-range MVS tasks. For example, SD-MVS~\cite{sdmvs2024} leveraged the Segment Anything Model (SAM)~\cite{k2023sam} for instance segmentation, enabling deformable propagation and matching within each segment, thereby achieving high-quality depth estimation with optimized memory use. MaGNet~\cite{magnet2022} combined monocular depth distribution estimation with multi-view geometry, achieving higher depth estimation accuracy by evaluating fewer depth candidates. NR-MVSNet~\cite{Li_NR_MVSNet} incorporated depth-normal consistency to guide depth candidate sampling, enhancing cost volume construction and depth inference. NVPA-MVSNet~\cite{nvpamvs2022} utilized normal priors for pixel visibility estimation, facilitating more reliable cost aggregation. Other methods \cite{virtualnor2019} \cite{normalstereo2020} \cite{normalocc2020} leveraged geometric priors through depth-normal consistency losses. More recently, GoMVS~\cite{wu2024gomvs} introduced a GCA operation that also leverages normal priors. By aggregating geometrically relevant matching results instead of using standard 3D convolution, the GCA operation enhanced close-range reconstruction accuracy.

In contrast to GoMVS~\cite{wu2024gomvs}, our approach further integrates monocular depth prior and improves the GCA for our remote sensing context. We extend the fixed neighboring positions to broader, geometry-aware neighbors to mitigate detail loss in long-range captures. Additionally, we incorporate neighboring feature similarity weights for more effective cost aggregation across expanded neighbor positions.

\subsection{Depth Range Sampling Strategies for Cascaded MVS}
Starting from~\cite{2020Cascade,2019Deep}, multi-stage MVS framework has become a standard pipeline, which constructs cascaded cost volumes for depth reconstruction in a coarse-to-fine manner. Depth range sampling plays a critical role in multi-stage cost volume formulation. Specifically, there are three key factors that determine the depth range sampling process: the depth range ($R$), the depth sampling interval (I), and the depth sampling number ($N$), where $R=I \times N$. Existing methods can be categorized based on which two factors are determined. Some methods, such as Cas-MVSNet~\cite{2020Cascade} and Uni-MVSNet~\cite{Peng_UniMVSNet}, predetermine both $I$ and $N$ using customized rules. Other methods, such as Vis-MVSNet~\cite{2020VisMVSNet} and UCSNet~\cite{2019Deep}, predict $R$ based on uncertainty-based metrics (e.g., entropy and standard deviation of matching scores) while keeping $N$ as hyperparameters. CVP-MVSNet~\cite{CVP_MVS_CVPR}, in contrast, calculates $I$ and $R$ by ensuring mean pixel offsets of 0.5 and 2 pixels in the image space.

Compared to existing depth range sampling strategies, our method leverages geometric cues to guide adaptive depth range prediction, thereby improving feature-matching effectiveness across diverse scene regions. For instance, in edge areas, the wider sampling range can provide more distinguishable information for accurate depth inference. In addition, our approach reduces dependence on pre-defined rules and uncertainty estimations, which may collapse in diverse and widely low-textured aerial imagery.

\section{Methodology}
Our approach incorporates readily available monocular geometric estimates from large-scale pre-trained models into multiple steps of the cascaded MVS pipeline, facilitating accurate aerial multi-view depth estimation. It comprises three key components: predicting adaptive depth range maps for cost volume construction (Sec.~\ref{sec:3_1}), performing normal-guided high-quality cost aggregation (Sec.~\ref{sec:3_2}), and applying normal-guided depth refinement (Sec.~\ref{sec:3_3}). Finally, we introduce the loss functions in Sec.~\ref{sec:3_4}. An overview of the framework is provided in Fig.~\ref{fig:arch}.

\noindent \textbf{Monocular Depth Cues.} We utilize the DepthAnythingV2 model~\cite{yang2024depth} to generate an initial depth map $D_m$ for each reference image. Notably, this monocular depth serves only as a relative cue, as it is inherently scale-ambiguous and may significantly deviate from the desired output. Based on this, we first perform a domain conversion operation to compute the depth-to-normal estimate $N_d$. Then, by performing depth-to-normal and monocular normal feature interactions, our method detects regions where monocular depth estimates are distorted and infers pixel-wise adaptive depth search range for cost volume construction, with minimized depth regression deviation. Consequently, the inferred depth search range maps for the first stage are generally larger than the predefined depth range value to promote discriminative feature matching in aerial stereo images.

\noindent \textbf{Monocular Normal Cues.} We employ the pre-trained Omnidata model~\cite{omnidata} to predict an initial normal map $N_m$ for each reference view. Normal priors provide valuable information about local depth variations, complementing depth cues. Additionally, they enhance cost aggregation by guiding the process based on geometry-aware 3D positions rather than the fixed 3D positions of 3D convolutions. The embedded geometric information in normal priors also aids in refining matching-based depth maps, further improving depth estimation accuracy.

\subsection{Depth Range Predictor}\label{sec:3_1}
The Depth Range Predictor (DRP) aims to generate a pixel-wise depth range map, $R_{pred}$, based on depth and normal estimates. Following the setting of existing cascaded architecture~\cite{wu2024gomvs}, DRP is applied three times for cost-volume construction. In the first stage, depth and normal maps are generated by pre-trained monocular geometric models~\cite{yang2024depth} \cite{omnidata}. For subsequent stages, the depth is taken from the regressed map of the previous stage, while the normal remains sourced from the monocular estimation.

As illustrated in Fig.~\ref{fig:arch}, the DRP is initially split into two branches: the depth and normal branches. The depth branch begins with domain conversion, scaling the monocular depth to a predefined range and then computing the normal map using the depth and camera intrinsics, following the closed-form solution presented in~\cite{qi2018geonet}. The obtained depth-to-normal map $N_d$ and monocular normal map $N_m$ are processed independently by a Monocular Feature Extraction (MFE) module, denoted as $f_{m}$. This module consists of a three-layer CNN, including $3 \times 3$ regular and depthwise separable convolution layers, as follows:
\begin{equation}
\left \{F_{d},F_{m}  \right \} =\left \{ f_{m}(N_d),f_{m}(N_m) \right \}.
\end{equation}

\begin{figure}[!t]
\begin{center}
\includegraphics[width=0.7\linewidth]{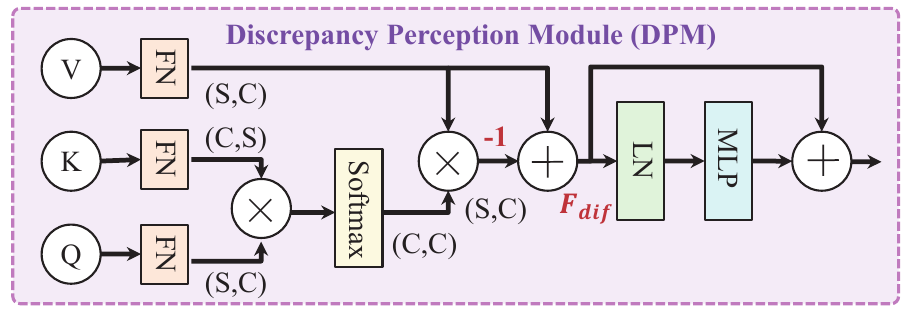}
\end{center}
\caption{Illustration of Discrepancy Perception Module. FN denotes the flattening operation; $\otimes$ represents matrix multiplication; and $\oplus$ indicates element-wise addition.}
\label{fig:dppm} 
\end{figure}

Then, how to fully exploit discrepancy and common information present in depth and normal features is a key factor for range map prediction. Motivated by the success of cross-attention mechanisms for feature alignment, we introduce a Discrepancy Perception Module (DPM), which extracts discrepancy features and predicts the range map. The DPM, shown in Fig.~\ref{fig:dppm}, operates in the channel dimension, per pixel, to learn the discrepancy feature. The monocular features $F_m$ and $F_d$ are segmented ($S_{eg}$) into $\left \{ Q, K, V \right \}$ components, as follows:
\begin{align}
Q_1,Q_2,...Q_S & = S_{eg}(F_m),\nonumber \\
K_1,K_2,...K_S & = S_{eg}(F_d),\\
V_1,V_2,...V_S & = S_{eg}(F_d), \nonumber
\end{align}
where $F_m$ and $F_d \in \mathbb{R}^{C  \times H/2 \times W/2}$, and $S = H/2 \times W/2$. Among them, $K$ and $V$ are derived from one set of features, while $Q$ is derived from the other. The similarity matrix between $Q$ and $K$ is computed, and the result is multiplied by $V$ to infer the shared information. Subsequently, the discrepancy feature $F_{dif}$ is obtained by subtracting the shared information from $V$.
This discrepancy feature is then enhanced through Layer Normalization (LN) and a Multi-Layer Perceptron (MLP) residual block to generate the predicted depth range map $R_{pred}$. In the DRP, DPM is applied twice, with the input feature orders exchanged:
\begin{align}
R_{pred}^1 &=f_{dpm}(Q_{1...S}^{F_m},K_{1...S}^{F_d},V_{1...S}^{F_d}), \nonumber \\
R_{pred}^2 &=f_{dpm}(Q_{1...S}^{F_d},K_{1...S}^{F_m},V_{1...S}^{F_m}), \\
R_{pred} &=R_{pred}^1+R_{pred}^2. \nonumber
\end{align}

Next, the predicted depth range map is used to construct the depth candidate volume, which then undergoes homography warping and cost volume construction, as in standard MVS pipeline~\cite{wu2024gomvs} \cite{2020PatchmatchNet}. Specifically, the depth candidate volume is centered uniformly around the current depth map and bounded by the predicted range map, as given by:
\begin{equation}\label{eq:1}
V_{d}^{k}=\left [ D_{current}^{k}-R_{pred}^{k},D_{current}^{k}+R_{pred}^{k}\right ],
\end{equation}
where $k=\left \{ 1,2,3 \right \} $ denotes the stage. As the stage progresses, the number of sampled depth candidates decreases, while the spatial resolution remains at half-resolution to optimize GPU memory usage. The homography warping operation is given by:
\begin{equation}\label{eq:2}
H_{i}^{d}=dK_iT_iT_{0}^{-1}K_{0}^{-1},
\end{equation}
where $i=\left \{1,2...N\right \}$ denotes the source view, and $i=0$ is the reference view. Here, $T_i$ and $K_i$  are the camera's extrinsic and intrinsic matrices, and $d$ represents per-pixel depth candidates. This operation projects deep features from the source views onto reference view, enabling feature similarity calculations, pixel-wise weight map estimation, and the construction of multi-view aggregated cost volume.

To ensure that the generated depth candidate volume includes the GT depth, the predicted depth range must be at least twice the monocular depth deviation. This constraint is enforced through a residual loss during training, as detailed in Section~\ref{sec:3_4}. By minimizing the loss, we optimize the adaptive range maps to improve cost volume construction and bring the smallest depth regression error. 

\subsection{Normal-guided Cost Aggregation}\label{sec:3_2}
We introduce the Normal-Guided Cost Aggregation (NCA) for high-quality cost aggregation in aerial MVS. First, we apply deformable learning to the monocular normal map to estimate per-pixel offsets, enabling the selection of geometrically relevant $K^{2}$ neighboring pixels. Meanwhile, the expanded receptive field resulting from deformable learning enhances the aggregation of more effective matching scores in aerial imagery with missed local details. Then, following GoMVS~\cite{wu2024gomvs}, for a given reference pixel $i$ and geometrically relevant adjacent pixel $j$, we align the depth candidates of pixel $j$ to the depth space of the reference pixel $i$ by calculating normal-converted depth ratios and then performing cost aggregation. Differently, we additionally calculate the feature similarity score between pixel $i$ and $j$ using reference feature map $F_0$ and incorporate this weight into the depth ratio calculation to prevent erroneous depth space alignment when $i$ and $j$ belong to disconnected planes with similar surface normals. 

Specifically, we model the 3D point relationship between pixels $i$ and $j$ using the monocular normal $N_m$ as follows:
\begin{equation}
N_m^T(X(u_i,v_i)-X(u_j,v_j))=0.
\end{equation}

The depth relationship between the reference pixel $i$ and adjacent pixel $j$ is given by:
\begin{equation}
\frac{d(u_i,v_i)}{d(u_j,v_j)} =\frac{N_m^T[\frac{u_i-c_x}{f_x} \frac{v_i-c_y}{f_y} 1 ]^T}{N_m^T[\frac{u_j-c_x}{f_x} \frac{v_j-c_y}{f_y} 1 ]^T},
\end{equation}
where $c_x$, $c_y$, $f_x$, and $f_y$ are the camera's intrinsic parameters. We define the depth ratio between pixel $i$ and $j$ as $r_{ji}=\frac{d(u_i,v_i)}{d(u_j,v_j)}\omega_f$. The term $\omega_f$ is the calculated feature similarity weight. 

Then, we perform geometrically consistent cost propagation, aggregating high-quality costs from geometry-relevant 3D positions to produce intermediate matching costs of size $K^{2}M \times D\times H/2 \times W/2$, which includes the matching results for $K^{2}$ geometrically relevant adjacent pixels, M feature channels, and D depth candidates. A $1 \times 1 \times K$  3D convolution is then applied for cost aggregation, yielding computational efficiency equivalent to a standard $K \times K \times K$ 3D convolution. For each pixel pair $i-j$, the depth ratio is determined by their respective depth candidates, while different neighboring pairs have distinct weights based on their feature similarities. We incorporate our NCA operator within the 3D U-Net architecture \cite{2018MVSNet} \cite{wu2024gomvs} to build our cost volume regularization network, which outputs a regularized probability volume $P \in \mathbb{R}^{D\times H/2 \times W/2}$. After regularization, we estimate the matching-based depth map $D_{pred}  \in \mathbb{R}^{1 \times H/2 \times W/2}$ in a regression manner, as follows: 
\begin{equation}
D_{pred}= \sum_{l=1}^{D}d_{l}P(d_{l}) .
\end{equation}

\subsection{Normal-guided Depth Refinement}\label{sec:3_3}
After three times cost volume construction and regularization, we produce the final matching-based depth estimation, $D_{pred}$, at half-resolution. To recover the full-resolution, we introduce a Normal-guided Depth Refinement (NDR) module that addresses the over-smoothed depth boundaries caused by the large receptive fields involved in regularization, yielding the final depth estimate $D_{ref} \in \mathbb{R}^{1 \times H \times W}$. Inspired by residual refinement strategies in PatchmatchNet~\cite{2020PatchmatchNet} and EG-MVSNet~\cite{egmvs2024}, we designed the NDR to enhance depth estimation accuracy. First, the coarse depth map $D_{pred}$ is upsampled using bilinear interpolation, resulting in the initial estimate $D_{up} \in \mathbb{R}^{1 \times H \times W}$. To mitigate depth scale influence, both $D_{up}$ and the monocular normal $N_{m}$ are pre-scaled to the $\left [ 0, 1 \right ]$ range. Feature maps $\hat{F_D}$ and $\hat{F_N}$ are independently extracted using 2D convolutional layers. These feature maps are concatenated and passed through additional 2D convolutional layers to generate the residual depth map, which is added back to $D_{up}$. The resulting refined depth map is then rescaled back to the original depth range. The formulation of our NDR is as follows:
\begin{small}
\begin{equation}
D_{ref}=(f_r(\left [f_{rd}((D_{pred})^{s}), f_{rn}((N_{m})^{s}) \right ] ) +(D_{pred})^{s})^{rs},
\end{equation}
\end{small} 
\noindent where $(\cdot)^s$ and $(\cdot)^{rs}$ represent the scale and rescale operations, and $f_{rd}$, $f_{rn}$, and $f_r$ denote the 2D convolutional layers.

\subsection{Optimization}\label{sec:3_4}
The total loss is defined as:
\begin{equation}
L_{total}=\sum_{k=1}^{3} (L_{r}^{k}+L_{d}^{k})+L_{ref},
\end{equation}
where $L_{r}^{k}$ supervises the predicted range maps for cost volume construction at $k$-th stage, $L_d^{k}$ supervises the matching-based depth maps at half-resolution, and $L_{ref}$ supervises the final refined depth map at full-resolution.

We apply the smooth L1 loss to predicted depth maps. For the predicted range maps, the ground-truth residual map $R_{gt}^{k}$ is computed by comparing the ground-truth depth map with the monocular depth estimation $\left ( k=1 \right )$ or the regressed depth map from the previous stage $\left ( k=2,3 \right )$. This residual is then used to construct the loss $L_r^{k}$ as follows:
\begin{footnotesize}
\begin{align}
R_{gt}^{k} & = \left\{\begin{matrix}
 \left | D_{gt}-D_{m} \right | & k = 1 \nonumber\\
 \left | D_{gt}-D_{pred}^{k-1} \right | &k = 2,3 \nonumber\\
\end{matrix}\right.\\
M_c^{k} & = \left\{\begin{matrix}
\Phi (D_m-R_{pred}^{k} < D_{gt} < D_m+R_{pred}^{k})  & k = 1 \\
\Phi (D_{pred}^{k-1}-R_{pred}^{k}<D_{gt}<D_{pred}^{k-1}+R_{pred}^{k}) & k = 2,3
\end{matrix}\right.\\
L_r^{k}& =\left\{\begin{matrix}
 (1-M_c^{k})(|D_{m}\pm R_{pred}^{k},D_{gt}|^2)+\left | R_{pred}^{k},R_{gt}^{k} \right | & k =1 \nonumber \\
 (1-M_c^{k})(|D_{pred}^{k-1}\pm R_{pred}^{k},D_{gt}|^2)+\left | R_{pred}^{k},R_{gt}^{k} \right | & k =2,3. \nonumber
\end{matrix}\right.
\end{align}
\end{footnotesize}

Here, $\Phi(\cdot)$ is an indicator function that returns 1 if the ground truth is within the predicted range, and 0 otherwise. The residual loss $L_r^{k}$ consists of two terms: the first penalizes cases where the generated depth candidates fail to encompass the GT depth, while the second prevents the predicted depth range from becoming excessively large.

\section{Experiment}

\subsection{Datasets and Evaluation Metrics}

To evaluate our method, we used three large-scale aerial MVS datasets: WHU~\cite{REDNet}, LuoJia-MVS~\cite{li2023hierarchical}, and M\"{u}nchen~\cite{haala2013munchen}. The WHU dataset, captured with a five-camera rig mounted on a UAV, spans an area of approximately 6.7 × 2.2 km² and includes urban regions, including dense urban buildings, open terrain, forests, and rivers. The LuoJia-MVS dataset offers broader land cover diversity, featuring cultivated fields, forests, urban, rural, industrial, and unused areas. Both the WHU and LuoJia-MVS datasets contain 4320 image pairs for training and 1360 pairs for testing, with each image having a resolution of 784 × 368 pixels and a spatial resolution of around 10 cm. The M\"{u}nchen dataset captured a metropolitan region and including 15 aerial images, each with dimensions of 7072 × 7776 pixels and a spatial resolution of about 10 cm.

We evaluate the quality of the estimated depth maps using three common metrics \cite{li2023hierarchical} \cite{REDNet} \cite{egmvs2024} \cite{mao2024sdl}: (1) Mean Absolute Error (MAE), which calculates the average L1-norm difference between the estimation and ground truth depth maps, considering only distances within 100 depth intervals to exclude extreme outliers; (2) The $< 0.6$ m, representing the percentage of pixels with an L1-norm error less than 0.6 m; and (3) The $< 3$-interval, which measures the percentage of pixels with an L1-norm error within three depth intervals. 

\begin{table*}[tp!]
\centering
\caption{Quantitative results on WHU~\cite{REDNet} and LuoJia-MVS~\cite{li2023hierarchical} datasets. Some results are obtained from EG-MVSNet \cite{egmvs2024}.}
\label{tab:whu_e} 
\scalebox{0.9}{
\begin{tabular}{c|c|ccc|ccc}
\hline\toprule
\multirow{2}{*}{\textbf{Number of Views}} & \multirow{2}{*}{\textbf{Method}} & \multicolumn{3}{c|}{\textbf{WHU}}     & \multicolumn{3}{c}{\textbf{LuoJia-MVS}}\\ \cline{3-8}
& & \textbf{MAE (cm)} & \textbf{\textless{}3-interval (\%)} & \textbf{\textless{}0.6 m (\%)} & \textbf{MAE (cm)} & \textbf{\textless{}3-interval (\%)} & \textbf{\textless{}0.6 m (\%)} \\
\midrule
& PatchmatchNet$^\diamondsuit$~\cite{2020PatchmatchNet} & 17.3     & 94.8& 96.5& 25.5& 87.2& 92.7\\
& R-MVSNet$^\diamondsuit$~\cite{RMVSNet} & 18.3     & 93.5& 95.3& 17.7& 93.5& 96.0\\
& RED-Net~\cite{REDNet} & 11.2     & 97.9& 98.1& 10.9& 96.9& 98.2\\
& Cas-MVSNet$^\diamondsuit$~\cite{2020Cascade} & 11.1     & 97.6& 97.7& 10.3& 97.1& 98.4\\
& HDC-MVSNet~\cite{li2023hierarchical} & 10.1     & 97.8& 97.9& 8.9& 97.8& {\color[HTML]{3531FF} \textbf{98.7}} \\
& SDL-MVS~\cite{mao2024sdl} & {\color[HTML]{3531FF} \textbf{9.5}} & {\color[HTML]{3531FF} \textbf{98.0}} & {\color[HTML]{3531FF} \textbf{98.4}} & {\color[HTML]{3531FF} \textbf{8.6}}  & {\color[HTML]{FE0000} \textbf{98.9}} & {\color[HTML]{FE0000} \textbf{98.9}} \\
& EG-MVSNet~\cite{egmvs2024} & 9.7& {\color[HTML]{3531FF} \textbf{98.0}} & 98.2& 8.7& 97.9& {\color[HTML]{FE0000} \textbf{98.9}} \\
& GoMVS$^\diamondsuit$~\cite{wu2024gomvs} & 12.4     & 96.5& 96.8& 13.1& 93.2& 93.4\\
\multirow{-9}{*}{Three-view}     & Ours   & {\color[HTML]{FE0000} \textbf{9.4}} & {\color[HTML]{FE0000} \textbf{98.4}} & {\color[HTML]{FE0000} \textbf{98.6}} & {\color[HTML]{FE0000} \textbf{8.2}}  & {\color[HTML]{3531FF} \textbf{98.1}} & {\color[HTML]{FE0000} \textbf{98.9}} \\
\midrule
   & PatchmatchNet$^\diamondsuit$~\cite{2020PatchmatchNet} & 16.0     & 95.0& 96.9& 28.3& 84.1& 90.4\\
   & R-MVSNet$^\diamondsuit$~\cite{RMVSNet} & 17.3     & 93.8& 95.4& 25.9& 86.7& 92.3\\
   & RED-Net~\cite{REDNet} & 10.4     & 97.9& 98.1& 15.6& 90.5& 94.9\\
   & Cas-MVSNet$^\diamondsuit$~\cite{2020Cascade} & 9.5& 97.8& 97.8& 14.1& 95.4& 97.9\\
   & HDC-MVSNet~\cite{li2023hierarchical} & 8.7& 98.0& 98.1& 12.1& 96.6& 98.3\\
   & SDL-MVS~\cite{mao2024sdl} & {\color[HTML]{262626} 8.2}& {\color[HTML]{FE0000} \textbf{98.9}} & {\color[HTML]{3531FF} \textbf{98.7}} & {\color[HTML]{262626} 11.8}& {\color[HTML]{FE0000} \textbf{98.2}} & {\color[HTML]{3531FF} \textbf{98.6}} \\
   & EG-MVSNet~\cite{egmvs2024} & {\color[HTML]{3531FF} \textbf{8.1}} & {\color[HTML]{3531FF} \textbf{98.7}} & {\color[HTML]{000000} 98.5}& {\color[HTML]{3531FF} \textbf{11.5}} & 96.9& 98.4\\
   & GoMVS$^\diamondsuit$~\cite{wu2024gomvs} & 11.6     & 97.6& 97.9& 14.4& 95.3& 96.1\\
\multirow{-9}{*}{Five-view}& Ours   & {\color[HTML]{FE0000} \textbf{7.8}} & {\color[HTML]{FE0000} \textbf{98.9}} & {\color[HTML]{FE0000} \textbf{99.1}} & {\color[HTML]{FE0000} \textbf{11.4}} & {\color[HTML]{3531FF} \textbf{97.3}} & {\color[HTML]{FE0000} \textbf{98.7}} \\ 
\bottomrule\hline
\end{tabular}}
\begin{tablenotes}
\footnotesize
\item{$^\diamondsuit$ indicates the methods proposed for close-range reconstruction. Others are specifically designed for large-scale aerial MVS task. The best and second-best results are in {\color[HTML]{FE0000}\textbf{red}} and {\color[HTML]{3531FF}\textbf{blue}}, respectively.}
\end{tablenotes}
\end{table*}

\subsection{Implementation Details}

We followed previous aerial MVS methods \cite{li2023hierarchical} \cite{REDNet} \cite{mao2024sdl} by training our models with three-view and five-view inputs on the training set, and evaluating their performance on the test sets of the WHU~\cite{REDNet} and LuoJia-MVS~\cite{li2023hierarchical} datasets. Additionally, we validated the generalization of models trained on the WHU dataset by testing them on the M\"{u}nchen dataset~\cite{haala2013munchen} without any finetuning.
All models were trained for 12 epochs to ensure a fair comparison, utilizing the Adam optimizer with parameters $\beta_1 = 0.9$ and $\beta_2 = 0.999$. The initial learning rate was set to 0.001 and was halved after the 6th and 8th epochs.
For the baseline model, GoMVS \cite{wu2024gomvs}, we maintained its original configuration, including the three-stage architecture with depth candidates set to $\left \{ 48,32,8 \right \} $, and the depth interval is halved after each stage. This is consistent with other aerial cascaded MVS networks \cite{2020Cascade} \cite{li2023hierarchical} \cite{mao2024sdl}. In contrast, for our model, which integrates the DRP, NCA, and NDR modules, we adjusted the cascade architecture to ensure that the computational memory usage remained comparable to the baseline. Specifically, we constructed three-stage cost volumes at half resolution and used the same depth candidates $\left \{ 48,32,8 \right \} $. The depth intervals were determined based on the predicted depth ranges and the predefined number of depth candidates.

\subsection{Benchmark Performance}

\begin{figure*}[!tp]
\begin{center}
\includegraphics[width=0.494\linewidth]{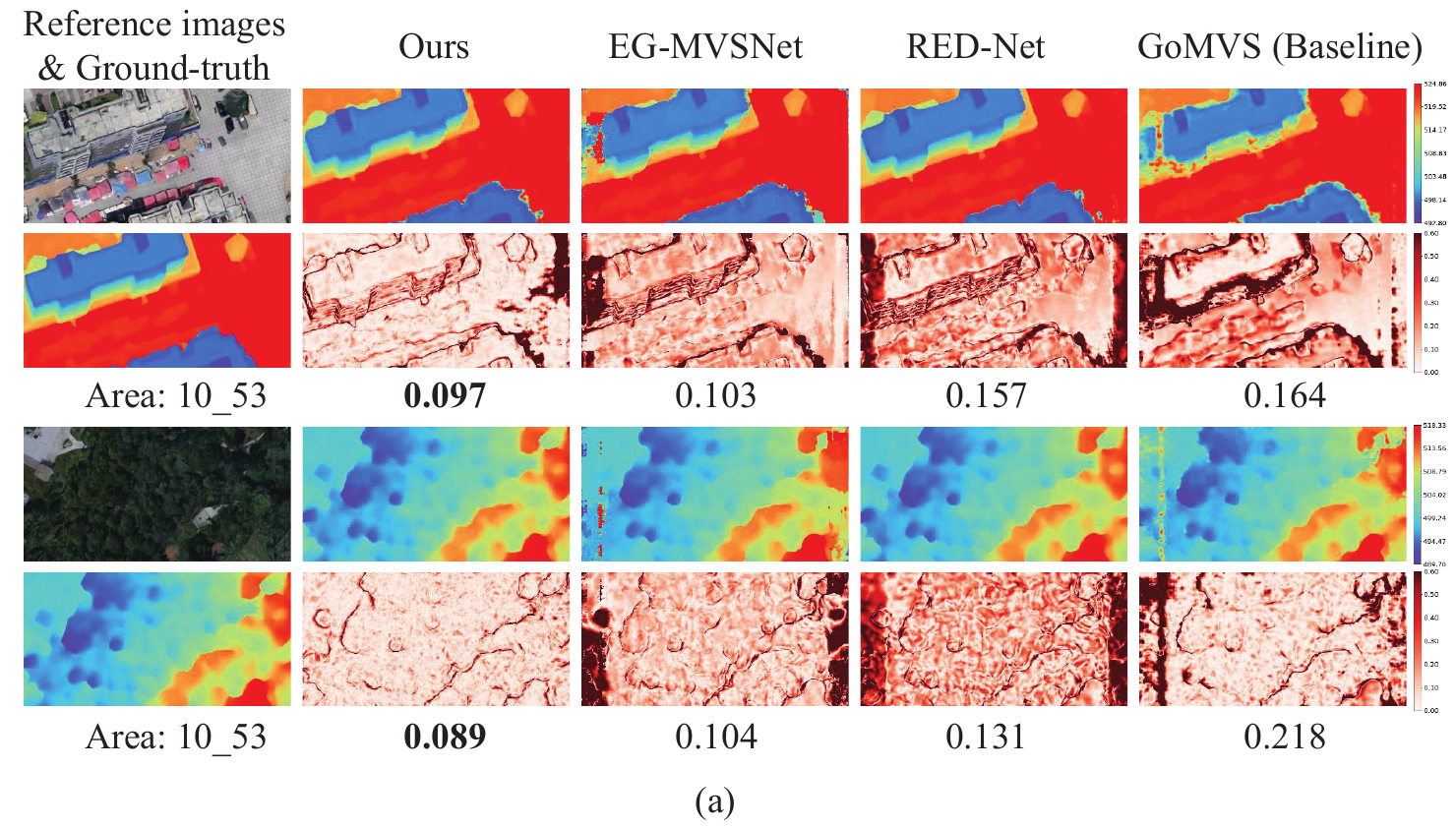}
\includegraphics[width=0.494\linewidth]{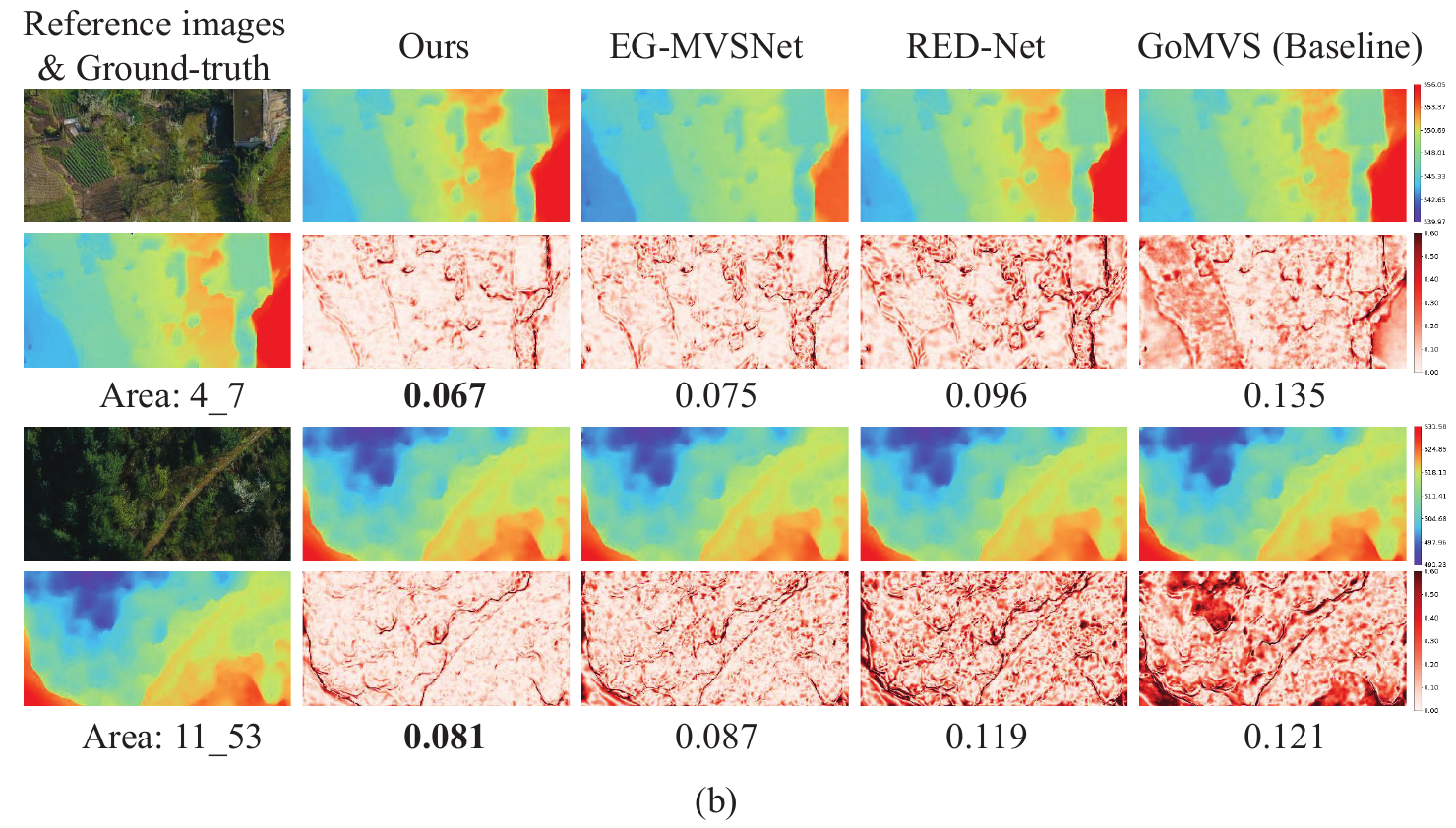}
\end{center}
\vspace*{-3mm}
\caption{Visualization comparison of depth estimation performance across different land types from the WHU (a) and LuoJia-MVS (b) datasets. The first and third rows present the estimated depth maps from EG-MVSNet, RED-Net, GoMVS, and our method. The second and fourth rows display the corresponding error maps, where higher red intensity signifies greater error.}
\label{fig:benchmark_depth_vis}
\end{figure*}

\textbf{Evaluation on WHU and LuoJia-MVS.}
We compare our approach to 8 typical deep MVS methods: PatchmatchNet~\cite{2020PatchmatchNet}, R-MVSNet~\cite{RMVSNet}, RED-Net~\cite{REDNet}, Cas-MVSNet~\cite{2020Cascade}, HDC-MVSNet~\cite{li2023hierarchical}, SDL-MVS~\cite{mao2024sdl}, EG-MVSNet~\cite{egmvs2024}, and GoMVS~\cite{wu2024gomvs}. Among these, GoMVS and EG-MVSNet are the most relevant and main competitors.
Quantitative evaluations on the WHU~\cite{REDNet} and LuoJia-MVS~\cite{li2023hierarchical} test sets are shown in Table~\ref{tab:whu_e}. Overall, our method achieves the best accuracy (lowest MAE) in both three- and five-view scenarios across both benchmarks. 

Three key observations are summarized: \textbf{1) Domain difference between close-range and aerial MVS:} Close-range MVS methods \cite{RMVSNet} \cite{2020PatchmatchNet} \cite{2020Cascade} \cite{wu2024gomvs} cannot be directly applied to aerial scenes, even when trained on aerial stereo images. For example, Cas-MVSNet \cite{2020Cascade} and GoMVS \cite{wu2024gomvs}, though effective for close-range reconstruction, yield comparable or inferior depth accuracy on aerial benchmarks. This performance gap mainly arises from differences in designed depth sampling processes. Close-range methods emphasize sampling within narrow and accurate depth ranges, typically realized by hand-crafted rules and uncertainty-based metrics. Accordingly, their initial sampling ranges are fixed and often too narrow for aerial settings. This leads to projected depth samples in the image being too close to provide distinguishable information, adversely impacting feature matching distinction and depth estimation accuracy. Our proposed DRP module addresses this domain gap by predicting an adaptive, enlarged depth range map to guide initial cost volume construction, fundamentally reducing optimization difficulty. Furthermore, our multi-stage depth sampling is driven by geometric cues, reducing reliance on pre-defined rules and uncertainty estimations, which may collapse in diverse and widely low-textured aerial imagery. As a result, our DRP can better adapt to the unique geometric characteristics of aerial MVS and achieve superior depth estimation accuracy.
\textbf{2) Integration of geometric cues enhances depth estimation:} Unlike EG-MVSNet~\cite{egmvs2024}, which incorporates geometric cues through jointly estimating depth and edge maps, our model embeds monocular geometric cues throughout the MVS pipeline. This strategy results in superior depth accuracy, as evidenced by our lower MAE and the improved visual detail in edge regions, as shown in Fig.~\ref{fig:benchmark_depth_vis}.
\textbf{3) Performance variations across datasets:} We compare the depth estimation performance in different land types on WHU and Luojia-MVS datasets in Fig.~\ref{fig:benchmark_depth_vis}. The dense building area of WHU introduces more severe occlusions, complicating accurate matching. Concerning the vegetated areas, EG-MVSNet and RED-Net perform better in the same land type on LuoJia-MVS but degrade on WHU. We attribute this to the advantage of diverse training data in LuoJia-MVS. In contrast, our approach and the baseline GoMVS, which leverage monocular normal cues, maintain better stability across both datasets.

\begin{table}[bp!]
\centering
\caption{Quantitative performance comparison on M\"{u}nchen dataset~\cite{haala2013munchen}. Some results are obtained from the EG-MVSNet~\cite{egmvs2024}. The best and second-best results are in {\color[HTML]{FE0000}\textbf{red}} and {\color[HTML]{3531FF}\textbf{blue}}, respectively.}
\label{tab:mun_e}
\resizebox{0.95\linewidth}{!}{
\begin{tabular}{c|c|ccc}
\hline\toprule
\textbf{Methods} & \textbf{Train Set} & \textbf{MAE (m)}& \textbf{\textless 3-interval (\%)}    & \textbf{\textless{}0.6 m (\%)} \\
\midrule
COLMAP~\cite{COLMAP_2016} & - & 0.5860& 73.36& 81.95\\
\midrule
RED-Net~\cite{REDNet}& WHU-3    & 0.5063& 80.67& 86.98\\
& WHU-5    & 0.5283& 80.40& 86.69\\
\midrule
EG-MVSNet~\cite{egmvs2024}  & WHU-3    & 0.4951& 82.16& 88.61\\
& WHU-5    & 0.4827& 82.56& {\color[HTML]{3531FF} \textbf{89.03}} \\
\midrule
Ours   & WHU-3    & {\color[HTML]{FE0000} \textbf{0.4719}} & {\color[HTML]{FE0000} \textbf{84.31}} & {\color[HTML]{FE0000} \textbf{89.05}} \\
& WHU-5    & {\color[HTML]{3531FF} \textbf{0.4812}} & {\color[HTML]{3531FF} \textbf{82.71}} & 88.96\\ 
\bottomrule\hline
\end{tabular}}
\end{table}

\begin{figure*}[!tp]
\begin{center}
\includegraphics[width=0.85\linewidth]{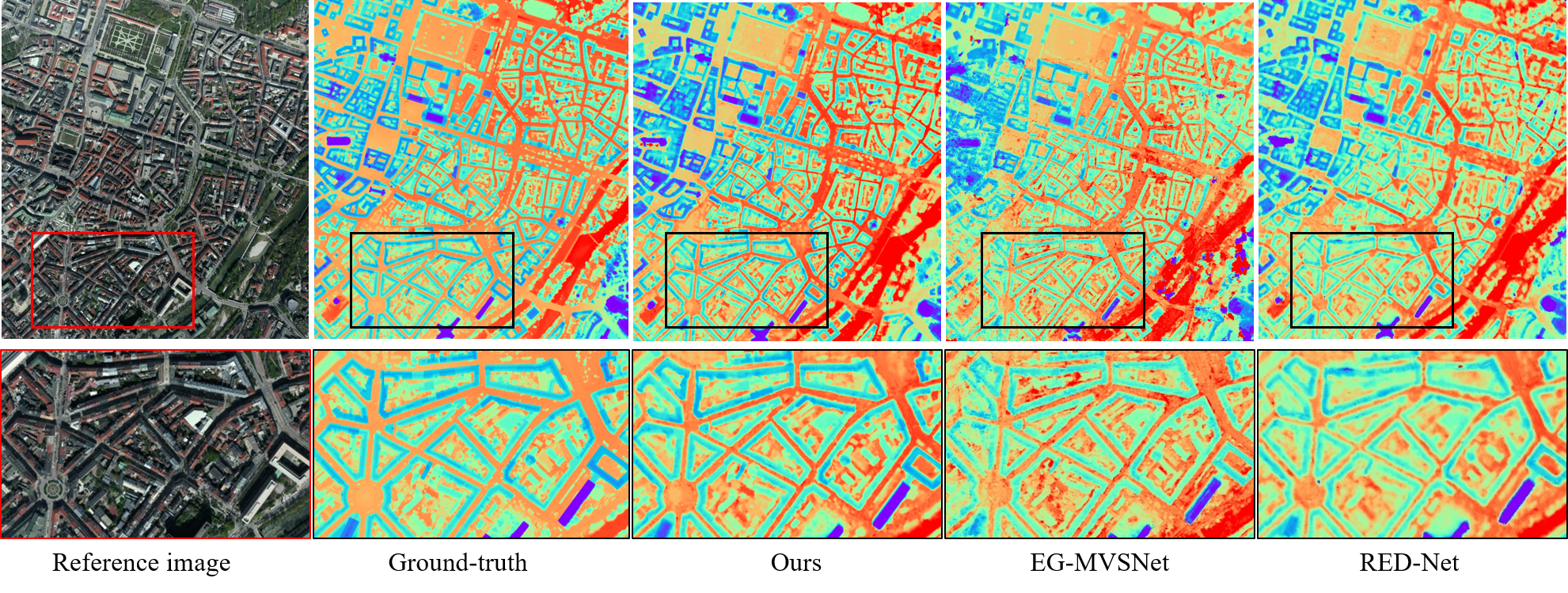}
\end{center}
\caption{Visualization comparison of depth estimation performance on M\"{u}nchen dataset~\cite{haala2013munchen}. All methods are trained on WHU-3 train set.}
\label{fig:mun_depth}
\end{figure*}

\textbf{Generalization on M\"{u}nchen dataset.}
We further evaluate our model on the M\"{u}nchen dataset~\cite{haala2013munchen}, with image resolution $3536 \times 3888$. The quantitative results are shown in Table~\ref{tab:mun_e}. Qualitative comparison with RED-Net and EG-MVSNet are shown in Fig.~\ref{fig:mun_depth}. Our method, trained on WHU-3, achieves the best performance across all evaluated metrics. Notably, it surpasses EG-MVSNet, which focuses on building scenes and incorporates manually annotated edge constraints during training. The superior performance of our method is primarily due to the adaptive range maps inferred for cascaded cost volume construction and the robust monocular normal cues integrated into the MVS pipeline.

In addition, we observe that the 5-view results are worse than the 3-view results, which contradicts the general trend in close-range MVS methods, where incorporating more views generally enhances reconstruction performance by providing richer geometric constraints and alleviating occlusion. We attribute this discrepancy to several aerial-specific factors, including dataset characteristics, distribution differences between the training and evaluation sets, and aspects of our model design. First, due to the top-down capture geometry and inherently low parallax in aerial imaging, adding more short baseline views sometimes introduces noisy or ambiguous matching cues rather than reinforcing correct correspondences. Second, the models in Table \ref{tab:mun_e} are trained on the WHU dataset, which covers diverse urban regions (e.g., sparse buildings, forests, rivers), where the occlusion of these flat areas is relatively limited and less complex. As a result, the trained model may be biased toward handling simpler occlusion configurations. When evaluated on the M\"{u}nchen dataset, which captures the metropolitan area, occlusions become more prevalent and intricate. As the number of input views increases from 3 to 5, the additional views may introduce shadows, occlusions, and photometric inconsistencies. These factors negatively impact feature matching and cost aggregation, especially in areas with repetitive textures, dense high-rise buildings, and reflective surfaces. Finally, our model employs the visibility-aware cost volume fusion module from the baseline GoMVS~\cite{wu2024gomvs}, composed of three 1×1 convolutional layers followed by sigmoid and max-pooling operations. While this design is lightweight and efficient, it may lack the capacity to robustly handle more complex occlusion patterns that emerge with additional views.

\begin{table}[t!]
\centering
\caption{Performance comparison on the WHU test set~\cite{REDNet}. The best and second-best results are in {\color[HTML]{FE0000}\textbf{red}} and {\color[HTML]{3531FF}\textbf{blue}}, respectively.}
\label{tab:gpu_e} 
\resizebox{0.9\linewidth}{!}{
\begin{tabular}{c|cccc}
\hline\toprule
\multirow{2}{*}{\textbf{Methods}} & \textbf{Number}   & \textbf{MAE}  & \textbf{GPU Mem.} & \textbf{Run Time}\\
  & \textbf{of Depth} & \textbf{(cm)} & \textbf{(MB)}    & \textbf{(s)}   \\
\midrule
R-MVSNet~\cite{RMVSNet} & 94 & 18.3 & \color[HTML]{FE0000}\textbf{1993} & 1.324   \\
RED-Net~\cite{REDNet} & 94 & 11.2 & 3199 & 0.814    \\
HDC-MVSNet~\cite{li2023hierarchical} & 48,32,8 & 10.1 & 7580 & - \\
EG-MVSNet~\cite{egmvs2024} & 94 & \color[HTML]{3531FF}\textbf{9.7} & 4567 & 1.234 \\
GoMVS~\cite{wu2024gomvs} & 48,32,8 & 12.4 & 2891 & \color[HTML]{3531FF}\textbf{0.477} \\
Ours & 48,32,8 & \color[HTML]{FE0000}\textbf{9.4} & \color[HTML]{3531FF}\textbf{2836} & \color[HTML]{FE0000}\textbf{0.466}\\ 
\bottomrule\hline
\end{tabular}}
\begin{tablenotes}
\footnotesize
\item{GPU memory consumption (GPU Mem.) and depth map inference time per view (Run Time) were measured using 3-view inputs, with each input image sized $768 \times 384$. The output depth map resolution is $192 \times 96$ for R-MVSNet and $768 \times 384$ for all other methods.}
\end{tablenotes}
\end{table}

\textbf{GPU Memory and Runtime.} We compare the efficiency of our method with leading MVS approaches in Table \ref{tab:gpu_e}, except for the newly proposed SDL-MVS~\cite{mao2024sdl}, without available code yet. Compared to both cascade-structured \cite{li2023hierarchical} \cite{egmvs2024} \cite{wu2024gomvs} and single-stage \cite{REDNet} \cite{RMVSNet} methods, our approach achieves the second-lowest GPU memory consumption, the shortest inference time, and the highest depth estimation accuracy. As reported in previous studies \cite{survey2021multi} \cite{survey2021deep}, the main memory consumption in learning-based MVS methods is the cost aggregation step. While R-MVSNet reduces memory usage by sequentially regularizing 2D cost maps via GRU and $4 \times$ downsampling cost volume, it compromises runtime. Our approach improves cost aggregation by replacing standard 3D convolution with the proposed NCA, customized aggregating geometry-relevant costs without additional memory overhead. Apart from that, our DRP and NDR modules collaboratively contribute towards achieving leading performance in both accuracy and efficiency.

\subsection{Ablation Study}

\begin{table}[tp!]
\centering
\caption{Ablation Study on LuoJia-MVS \cite{li2023hierarchical} datasets. The best and second-best results are in {\color[HTML]{FE0000}\textbf{red}} and {\color[HTML]{3531FF}\textbf{blue}}, respectively.}
\label{tab:abl_sum}
\resizebox{\linewidth}{!}{
\begin{tabular}{cc|ccc}
\hline\toprule
\multicolumn{2}{c|}{\multirow{2}{*}{\textbf{Ablations}}}    & \textbf{MAE}  & \textbf{\textless{}3-interval} & \textbf{\textless{}0.6 m} \\
\multicolumn{2}{c|}{}     & \textbf{(cm)} & \textbf{(\%)} & \textbf{(\%)}   \\ \midrule
\multicolumn{1}{c|}{\multirow{6}{*}{\begin{tabular}[c]{@{}c@{}}Depth \\ Volume\\  Generation\end{tabular}}}  & w/o DRP    & 10.1& 96.8& 97.1     \\
\multicolumn{1}{c|}{}   & w/o DRP, 2 × Pre. Range     & 9.6 & 97.3& 97.4     \\
\multicolumn{1}{c|}{}   & w/o DRP, 3 × Pre. Range     & 14.2& 94.6& 96.8     \\
\multicolumn{1}{c|}{}   & D + N DRP  & 9.8 & 97.3& 97.5     \\
\multicolumn{1}{c|}{}   & N2D + D DRP& {\color[HTML]{3531FF}\textbf{9.3}} & {\color[HTML]{3531FF}\textbf{97.6}}& {\color[HTML]{3531FF}\textbf{97.9}}     \\
\multicolumn{1}{c|}{}   & D2N + N DRP (Ours)& {\color[HTML]{FE0000}\textbf{8.2}}  & {\color[HTML]{FE0000}\textbf{98.1}} & {\color[HTML]{FE0000}\textbf{98.9}}   \\
\midrule
\multicolumn{1}{c|}{\multirow{4}{*}{\begin{tabular}[c]{@{}c@{}}Cost \\ Aggregation\end{tabular}}} & Standard Conv3D (3 × 3 × 3) & 9.8 & 97.3& {\color[HTML]{3531FF}\textbf{97.6}}     \\
\multicolumn{1}{c|}{}   & Standard Conv3D (5 × 3 × 3) & 9.9 & 97.2& {\color[HTML]{3531FF}\textbf{97.6}}     \\
\multicolumn{1}{c|}{}   & GCA (3 × 3 × 3)   & {\color[HTML]{3531FF}\textbf{9.7}} & {\color[HTML]{3531FF}\textbf{97.5}}& 97.5     \\
\multicolumn{1}{c|}{}   & NCA (Ours) & {\color[HTML]{FE0000}\textbf{8.2}}  & {\color[HTML]{FE0000}\textbf{98.1}} & {\color[HTML]{FE0000}\textbf{98.9}}   \\
\midrule
\multicolumn{1}{c|}{\multirow{4}{*}{\begin{tabular}[c]{@{}c@{}}Depth \\ Upsampling \\ \& Refinement\end{tabular}}} & Bilinear Upsampling   & 8.9 & 97.3& 98.0     \\
\multicolumn{1}{c|}{}   & Depth Ref. of MSVNet  & 8.7 &  {\color[HTML]{3531FF}\textbf{97.9}}& {\color[HTML]{3531FF}\textbf{98.3}}     \\
\multicolumn{1}{c|}{}   & Depth Ref. of PatchmatchNet & {\color[HTML]{3531FF}\textbf{8.6}} & 97.6& {\color[HTML]{3531FF}\textbf{98.3}}\\
\multicolumn{1}{c|}{}& NDR (Ours) & {\color[HTML]{FE0000}\textbf{8.2}}  & {\color[HTML]{FE0000}\textbf{98.1}} & {\color[HTML]{FE0000}\textbf{98.9}}   \\
\bottomrule\hline
\end{tabular}}
\end{table}

\begin{figure*}[!t]
\begin{center}
\includegraphics[width=0.75\linewidth]{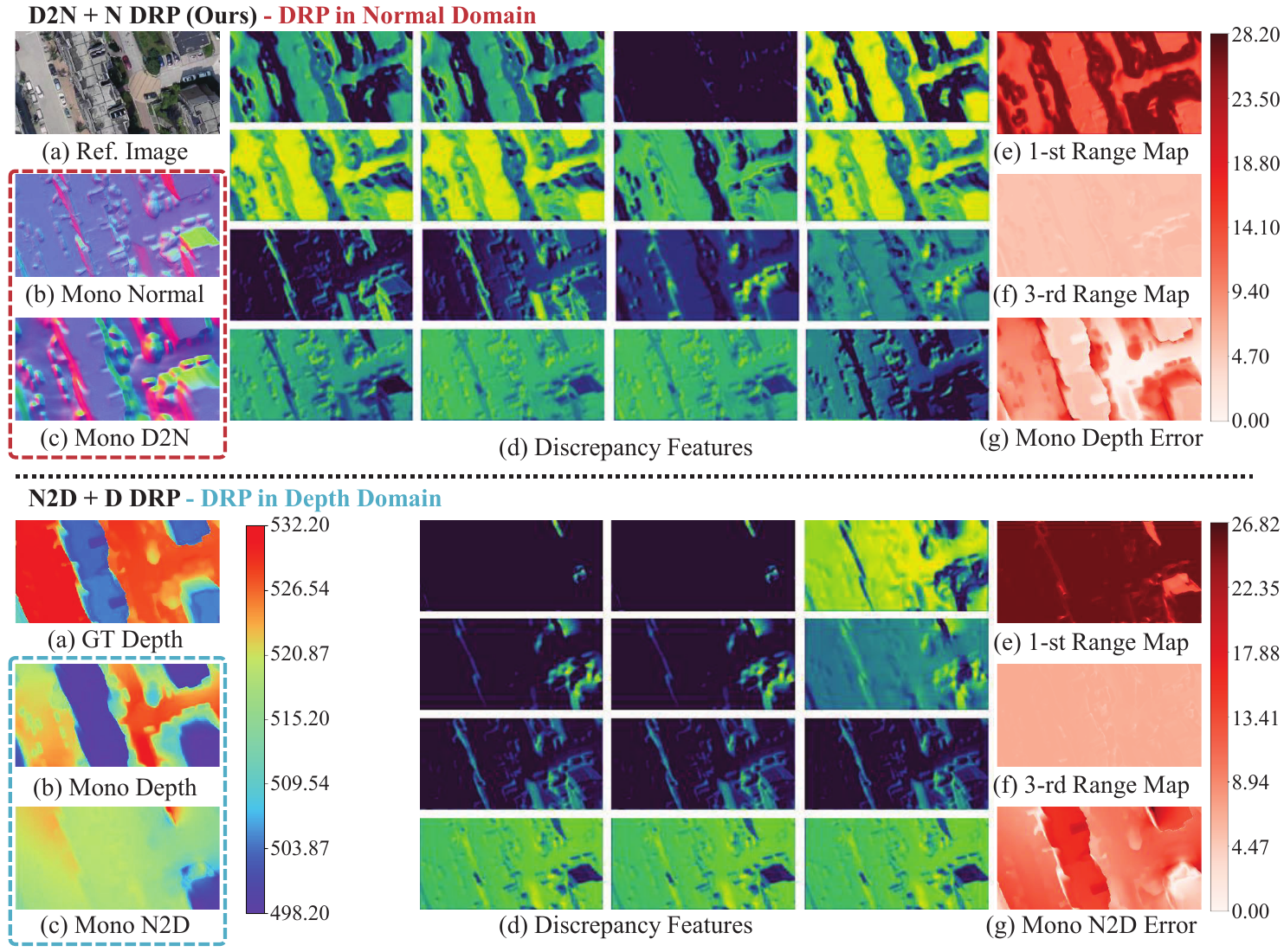}
\end{center}
\caption{Visualization of the intermediate results of two DRP designs. (b-c): The inputs of the MFE module. (d): Partial channels of the discrepancy feature $F_{dif}$ generated by the DPM. (e-f): The range maps predicted by the first and last stages.}
\label{fig:abl_ddp} 
\end{figure*}

\begin{figure*}[!h]
\centering
\includegraphics[width=0.45\linewidth]{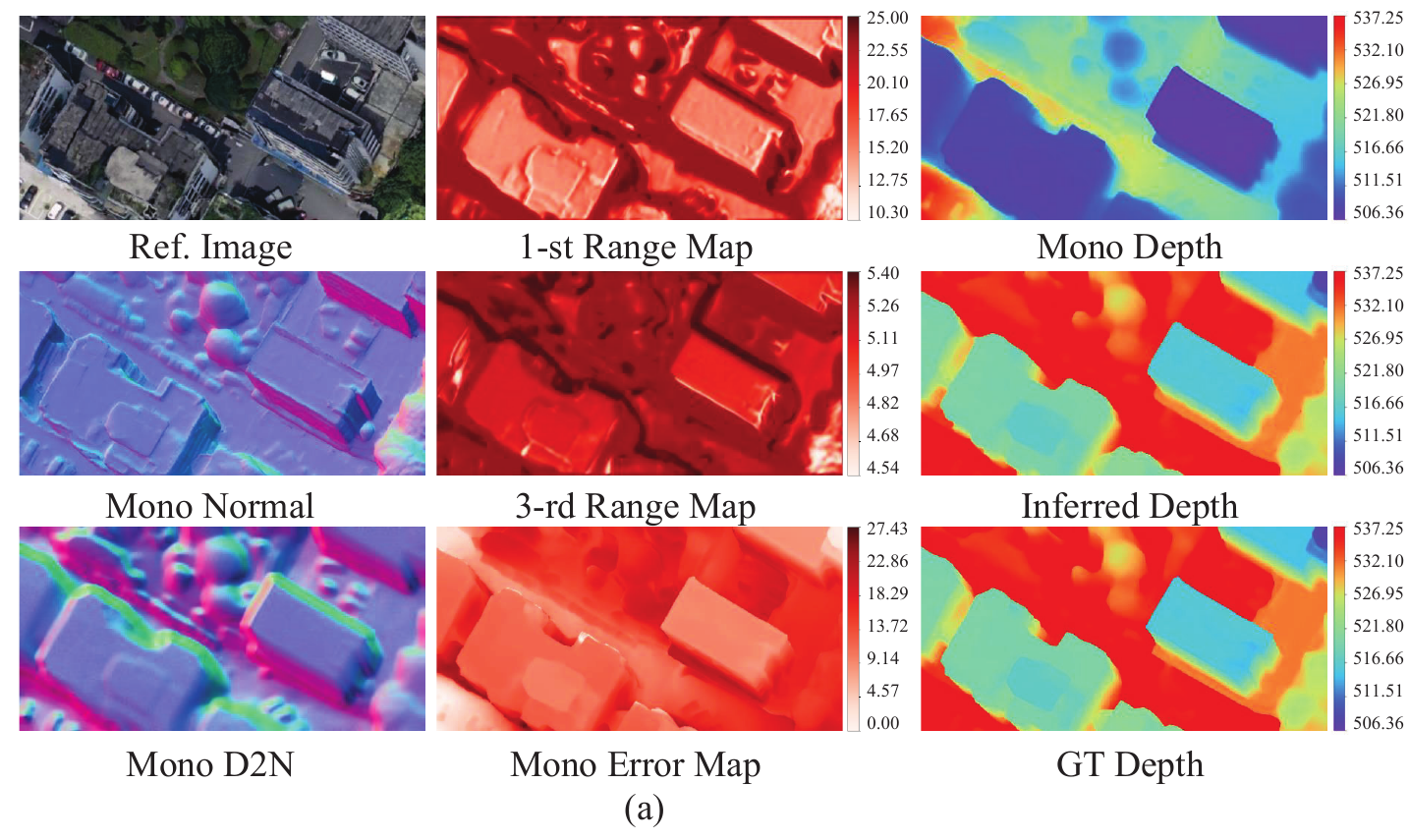}
\includegraphics[width=0.45\linewidth]{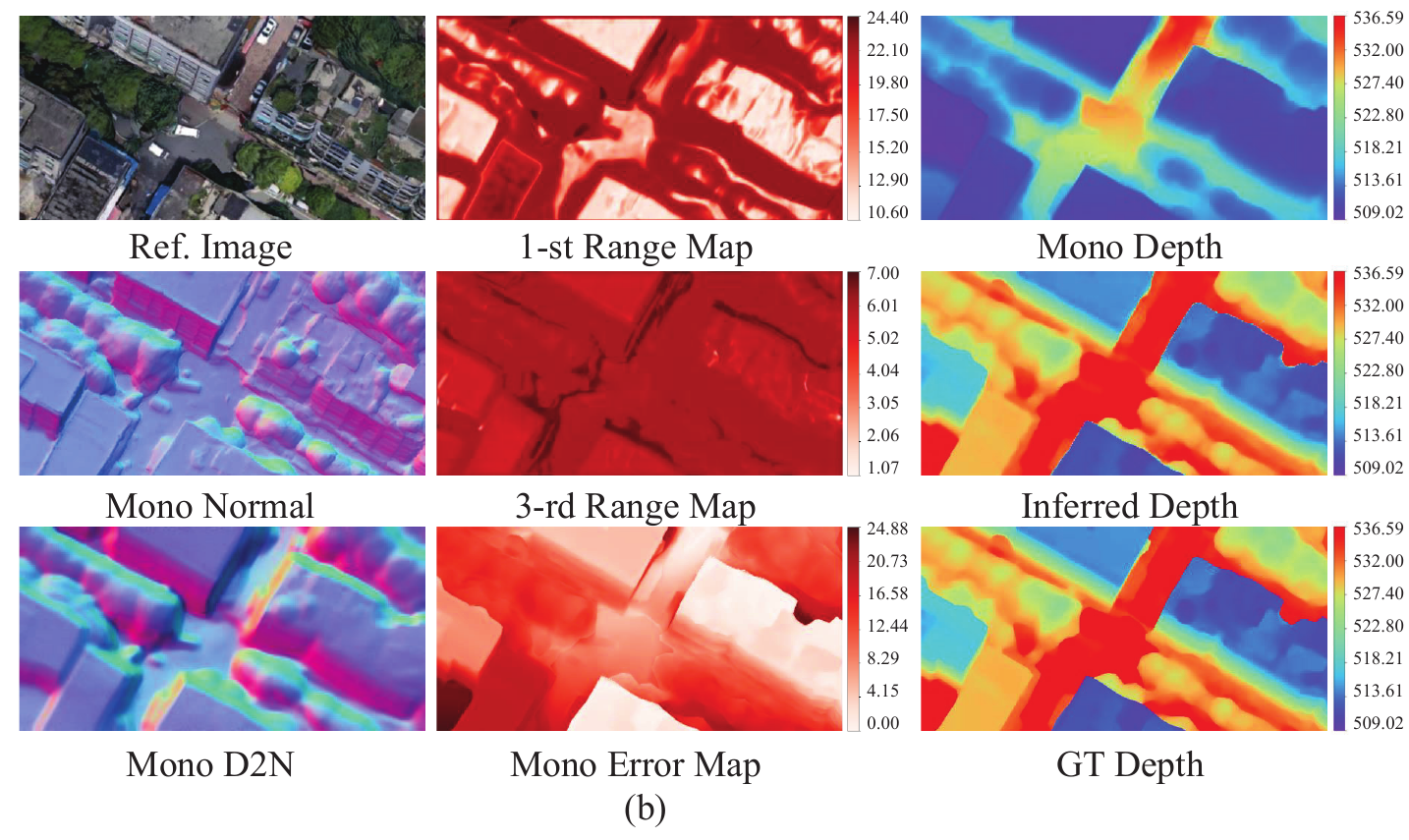}
\includegraphics[width=0.45\linewidth]{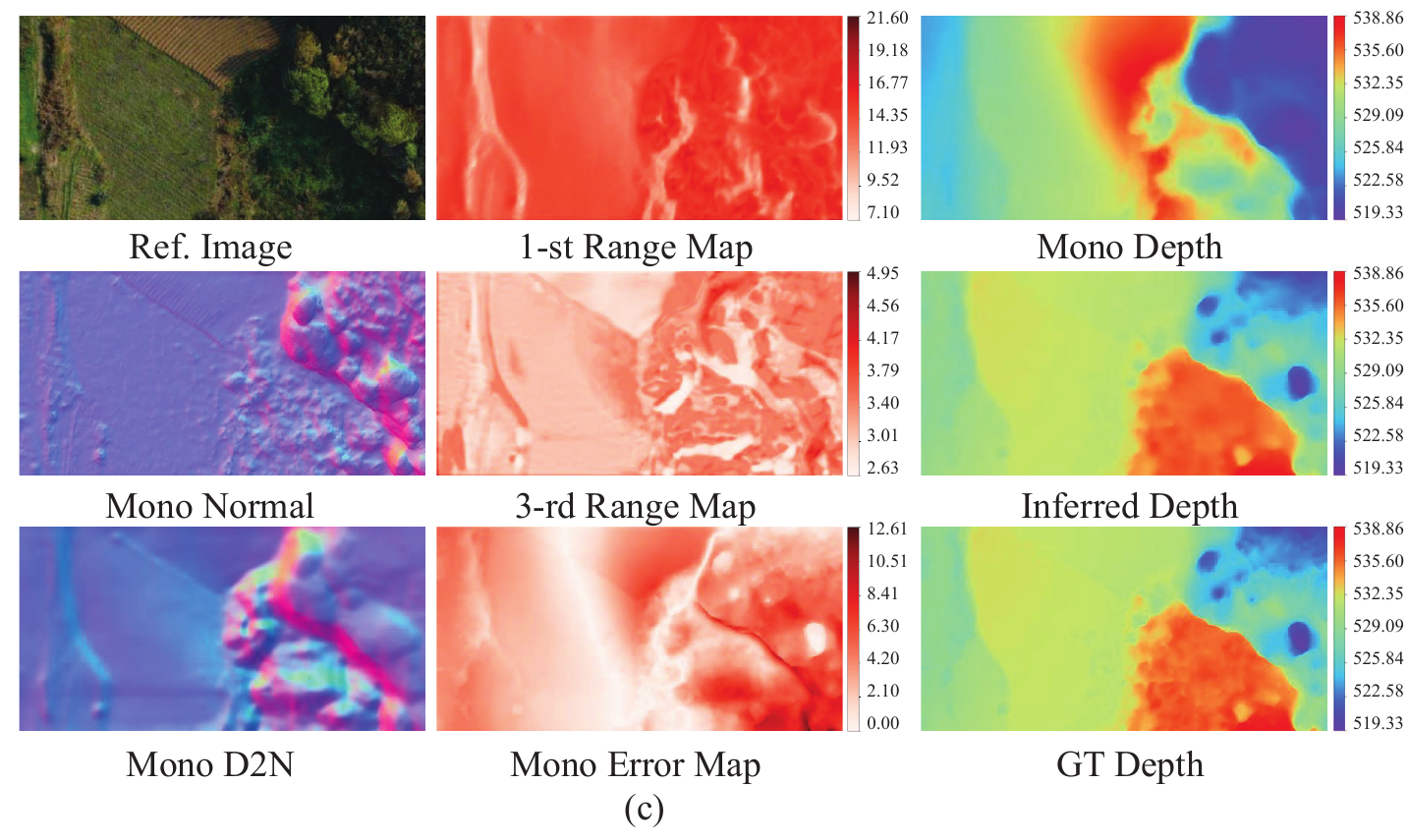}
\includegraphics[width=0.45\linewidth]{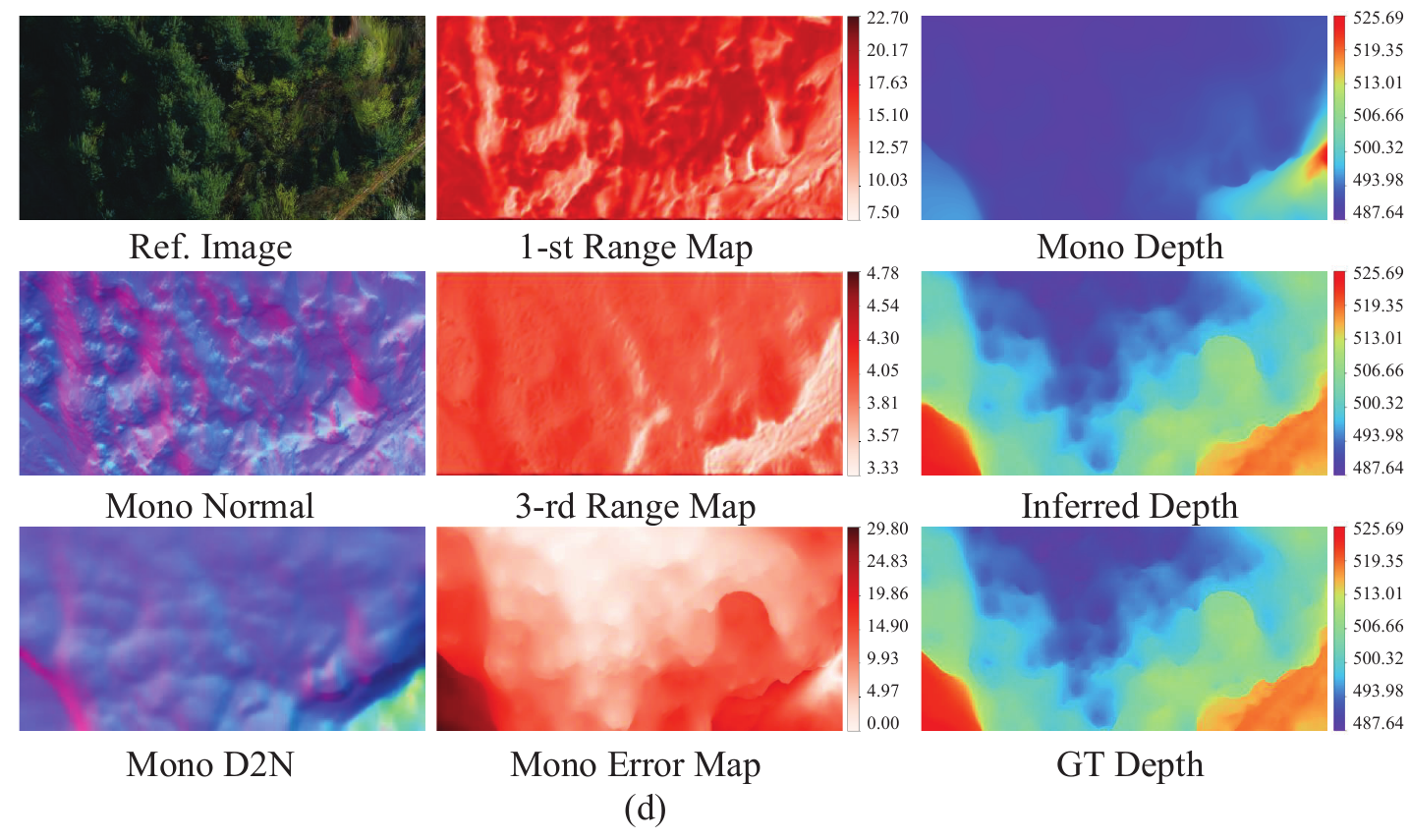}
\centering
\caption{Visualization of predicted range maps across multiple stages. (a) and (b) illustrate examples from the WHU dataset~\cite{REDNet}, while (c) and (d) present examples from the LuoJia-MVS dataset~\cite{li2023hierarchical}. Each example includes the reference image, monocular normal and depth predictions, depth-to-normal estimates, predicted range maps for the first and last stages, the final inferred depth map, and the ground truth depth. As the stages advance, the predicted range maps shift from an initially broader range to a narrower one. The final range values are comparable to those produced by cascaded MVS methods.}
\label{fig:drp_vis} 
\end{figure*}

\textbf{Effectiveness of DRP.} We evaluate different patterns of depth volume generation for our three-stage cost volume construction. The results are summarized in Table~\ref{tab:abl_sum}, and some intermediate outcomes are shown in Fig.~\ref{fig:abl_ddp}. 
First, rows 1-3 of Table~\ref{tab:abl_sum} present the impact of different predefined depth ranges. The first model (w/o DRP) uses dataset-provided ranges, while the second and third models expand these ranges to twice and three times their original sizes. The results indicate that doubling the range improves accuracy, but tripling it reduces performance. We assume this is because an over-large depth search range is adverse to precise depth regression in the cascaded architecture.

Second, rows 4-6 of Table~\ref{tab:abl_sum} compare different DRP designs. In the “D + N DRP” configuration, the depth and normal branches skip domain conversion, directly using normalized depth and normal estimates for monocular feature extraction and cross-attention feature interaction to predict range maps. In the “N2D + D DRP” design, domain conversion is applied to the normal branch via normal integration~\cite{normalintegration}. The produced height map (first stage), matching-based depths (later stages), and monocular depth are all normalized for three rounds of range map prediction. Here, range maps are predicted from the depth domain using two different estimates. The “D2N + N DRP” design, as described in the method section, applies domain conversion to the depth branch, predicting range maps from the normal domain.
Comparing the three designs, we observe that adaptive depth range prediction from the normal domain yields the best performance. This is because depth-to-normal conversion introduces fewer errors than normal-to-depth manner, particularly in areas with significant height variation. Our design better captures monocular depth deviations. As shown in Fig.~\ref{fig:abl_ddp}(c), the resulting range map of the first stage vaguely exhibits a positive correlation trend with the monocular depth error map, whereas the “N2D + D DRP” model fails to exhibit this relationship.

Fig.~\ref{fig:drp_vis} illustrates how the predicted range map is updated throughout the pipeline. In the first stage, the expanded range maps are generated based on exploring the divergence of monocular depth and normal features. After two rounds of multi-view matching, the predicted range maps in the final stage are significantly narrowed, ensuring that the spacing between depth samples is diminutive for precise depth regression. Additionally, we observe that the inferred range maps tend to exhibit larger range values in regions with significant normal variation across stages. We attribute this to the goal of facilitating precise depth estimation in edge regions.

\textbf{Effectiveness of NCA.}
We compare various cost aggregation techniques in Table~\ref{tab:abl_sum}, using the consistent 3D U-Net architecture for cost volume regularization. For standard 3D convolution, we tested different receptive fields in the depth dimension. GCA denotes the Geometrically Consistent Aggregation operation from GoMVS~\cite{wu2024gomvs}. Results show that expanding receptive fields in the depth dimension does not lead to improved accuracy. While GCA, effective in close-range MVS, offers only marginal gains in the aerial context. In contrast, our proposed NCA outperforms the standard 3D convolution by 1.3\% in the pixel ratio within 0.6 m bias. We attribute this improvement to the aggregation of expanded, geometry-relevant neighboring costs and the incorporation of feature similarities, which enhances the quality of cost aggregation, particularly for low-detail aerial stereo images.

\textbf{Effectiveness of NDR.}
In contrast to basic bilinear upsampling, MVSNet~\cite{2018MVSNet} and PatchmatchNet~\cite{2020PatchmatchNet} have integrated RGB features to enhance matching-based depth estimation while simultaneously performing upsampling. A key distinction between our approach and previous works \cite{2018MVSNet} \cite{2020PatchmatchNet} is the incorporation of monocular normal features as guidance.
As illustrated in Table \ref{tab:abl_sum}, the integration of both RGB and normal features with learnable parameters leads to an improvement in baseline performance. However, our normal-guided depth refinement yields the most favorable results. This suggests that the normal guidance signal is more direct and stable for our aerial images captured from long distances, highlighting the effectiveness of our approach.

\begin{table}[h!]
\centering
\caption{Impact of Monocular Depth and Normal Priors. Ablation results on the LuoJia-MVS dataset~\cite{li2023hierarchical} using three input views per sample.}
\label{tab:ness_mono}
\resizebox{0.85\linewidth}{!}{
\begin{tabular}{c|ccc}
\hline\toprule
\textbf{Method} & \textbf{MAE (cm)} & \textbf{\textless 3-interval (\%)}   & \textbf{\textless 0.6m (\%)} \\ \midrule
Ours (w/o D)    & 10.2 & 97.2 & 97.5 \\
Ours (w/o N)    & 13.6 & 93.5 & 93.3 \\ \midrule
Ours & 8.2 & 98.1 & 98.9 \\\bottomrule\hline
\end{tabular}}
\end{table}

\begin{table*}[tp!]
\centering
\caption{Generalization of ADR-MVS w.r.t Monocular Priors.}
\label{tab:gen_mono}
\vspace*{-1mm}
\resizebox{0.95\linewidth}{!}{
\begin{tabular}{cc|ccc|ccc|ccc}
\hline\toprule
\multicolumn{2}{c|}{\multirow{2}{*}{\textbf{Models}}} & \multicolumn{3}{c|}{\textbf{Luojia-MVS Test Set}} & \multicolumn{3}{c|}{\textbf{Memory Consumption (MB)}}    & \multicolumn{3}{c}{\textbf{Run Time (s)}}\\ \cline{3-11} 
\multicolumn{2}{c|}{}& \textbf{MAE (cm)} & \textbf{\textless{}3-interval (\%)} & \textbf{\textless{}0.6 m (\%)} & \textbf{Prior Step} & \textbf{MVS Step} & \textbf{Total} & \textbf{Prior Step} & \textbf{MVS Step} & \textbf{Total} \\
\midrule
ID-0 & Baseline~\cite{wu2024gomvs} (Omni~\cite{omnidata})    & 13.1    & 93.2    & 93.4& 0  & 2891    & 2891 & 0  & 0.477   & 0.477\\
ID-1 & ADR-MVS (DAv2~\cite{yang2024depth} + Omni~\cite{omnidata})     & 8.2     & 98.1    & 98.9& 625& 2836    & 3461 & 0.051     & 0.415   & 0.466\\
ID-2 & ADR-MVS (DAv2~\cite{yang2024depth} + Lotus~\cite{he2024lotus})     & 9.1     & 97.1    & 97.3& 878& 2836    & 3714 & 0.163     & 0.416   & 0.579\\
ID-3 & ADR-MVS (DepthPro~\cite{DepthPro2024} + Omni~\cite{omnidata}) & 8.9     & 97.7    & 98.2& 1976& 2836    & 4812 & 0.422     & 0.416   & 0.838\\
ID-4 & ADR-MVS (DepthPro~\cite{DepthPro2024} + Lotus~\cite{he2024lotus}) & 9.4     & 97.2    & 97.7& 1976& 2836    & 4812 & 0.534     & 0.416   & 0.95 \\
ID-5 & ADR-MVS (Lotus~\cite{he2024lotus} + Omni~\cite{omnidata}) & 10.8 & 96.8 & 97.2 & 878 & 2836 & 3714 & 0.233 & 0.416 & 0.649\\
ID-6 & ADR-MVS (Lotus~\cite{he2024lotus} + Lotus~\cite{he2024lotus}) & 9.7 & 97.5 & 97.6 & 890 & 2836 & 3726 & 0.345     & 0.416   & 0.761\\ \bottomrule\hline
\end{tabular}}
\begin{tablenotes}
\footnotesize
\item{Following the baseline GoMVS~\cite{wu2024gomvs}, we divided images into multiple overlapping patches to generate high-resolution normal cues using Omnidata~\cite{omnidata}. The pre-processing of Omnidata’s monocular normal cues was conducted without recording memory consumption and run-time.}
\end{tablenotes}

\end{table*}
\begin{figure*}[!ht]
\begin{center}
\includegraphics[width=0.8\linewidth]{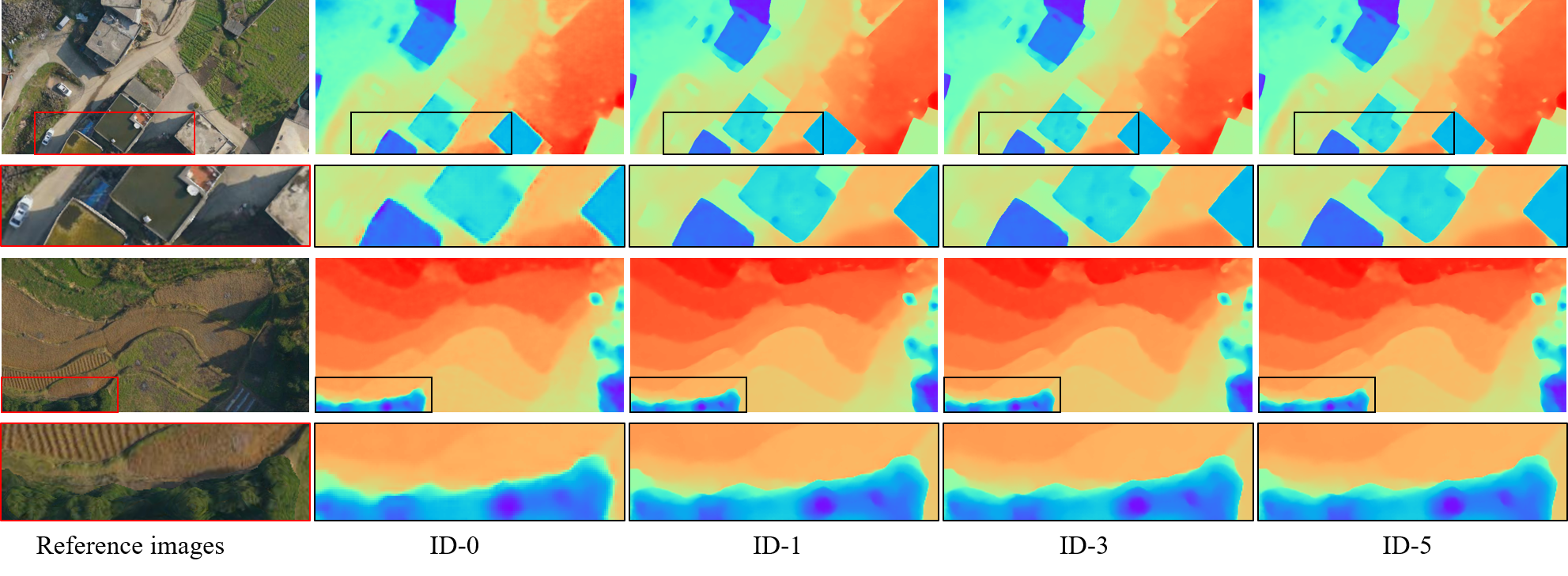}
\end{center}
\vspace*{-3mm}
\caption{Qualitative results of ADR-MVS with different monocular priors on LuoJia-MVS 109\_7 and 109\_21 areas. Model IDs correspond to Table~\ref{tab:gen_mono}.}
\label{fig:dif_mono_vis}
\end{figure*}

\begin{figure*}[!h]
\begin{center}
\includegraphics[width=0.7\linewidth]{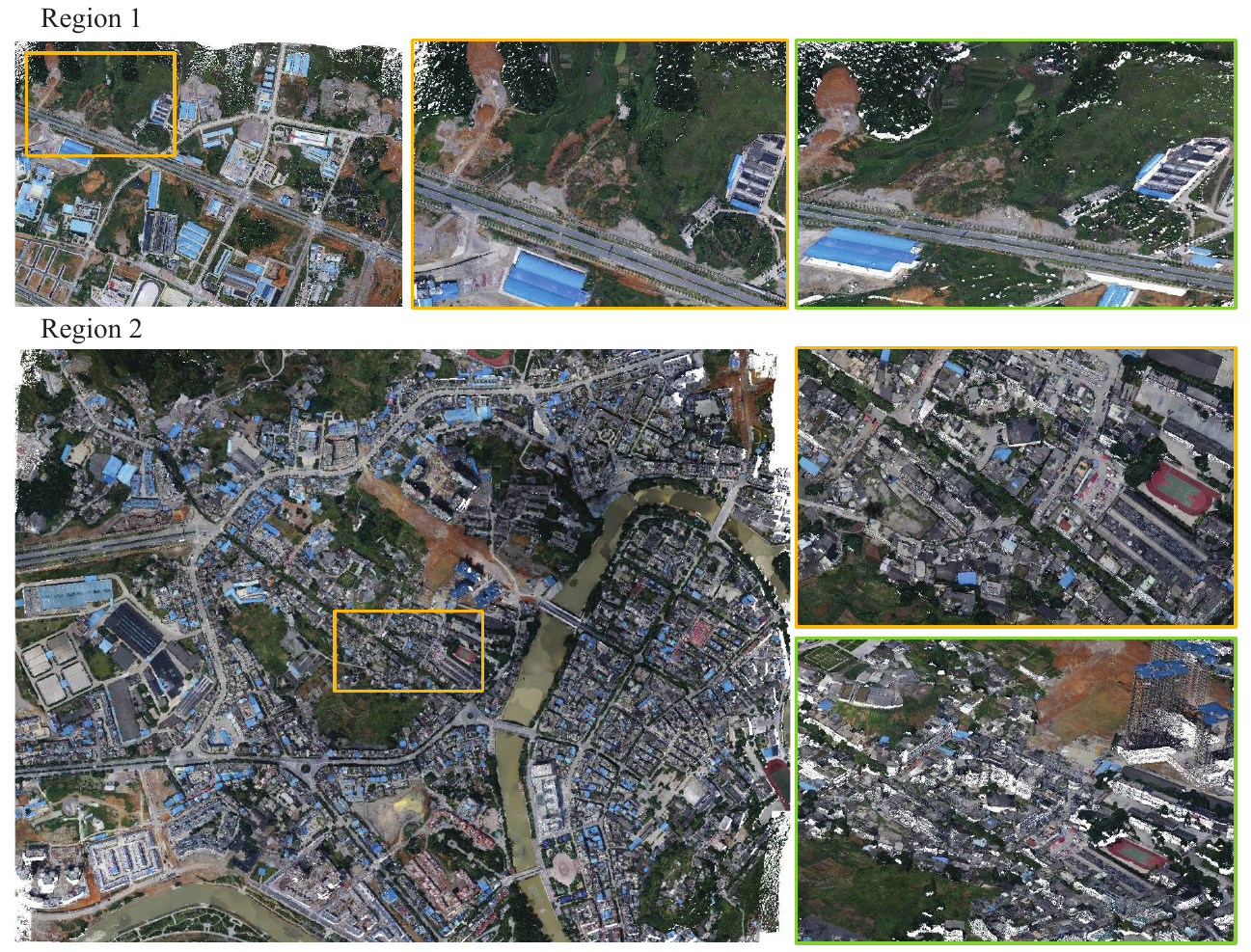}
\includegraphics[width=0.7\linewidth]{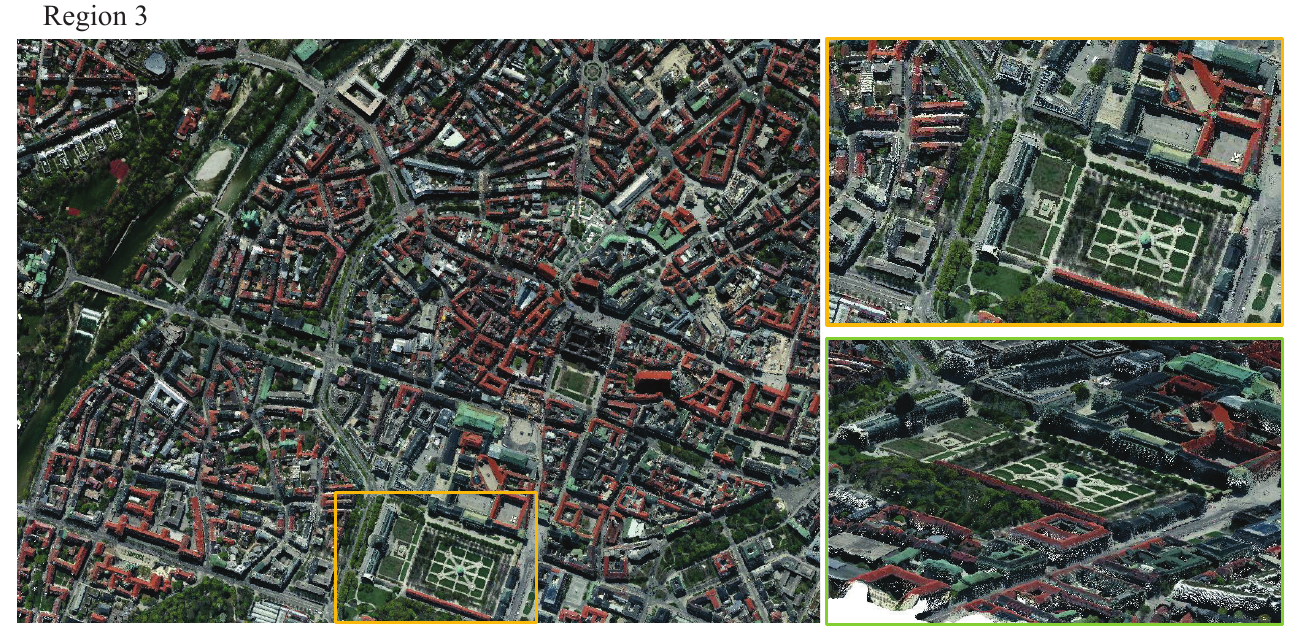}
\end{center}
\vspace*{-3mm}
\caption{\textbf{Visualization of Point Cloud Reconstructions.} Images with {\color[HTML]{F8B80A}\textbf{orange}} borders represent enlarged sections from the left scenes, while images with {\color[HTML]{83CD31}\textbf{green}} borders depict views from different perspectives.}
\label{fig:pc_vis}
\end{figure*}

\textbf{Necessity of Monocular Geometric Priors.}
To assess the necessity of monocular geometric priors in aerial MVS, we conducted ablation studies by training two ADR-MVS variants without either the monocular depth or normal priors and evaluated them on the LuoJia-MVS dataset~\cite{li2023hierarchical} (see Table~\ref{tab:ness_mono}). In “Ours (w/o D)”, the initial depth is randomly initialized within the predefined ranges and combined with monocular normals~\cite{omnidata} for range prediction and coarse-stage estimation. In “Ours (w/o N)”, normals are derived from monocular depth~\cite{yang2024depth} in the first stage and recomputed from estimated depths in subsequent stages in closed-form solution~\cite{qi2018geonet}.
Experimental results indicate that the absence of either prior leads to a noticeable decline in reconstruction accuracy, highlighting the complementary roles of both priors. The monocular depth prior provides initial estimates and strong geometric cues for adaptive depth range prediction, effectively guiding the MVS pipeline, particularly in the early stage where multi-view matching is ambiguous or underconstrained. The monocular normal prior plays vital roles in depth range prediction, cost aggregation, and refinement. Replacing it with depth-computed normals significantly degrades performance, as it depends on unreliable depth in challenging matching regions. In contrast, monocular normals, though lacking multi-view consistency, remain stable in ambiguous areas. This nice property effectively complements matching-based estimation.
Overall, the experiment confirms that monocular depth and normal priors are both essential components in our framework. They provide reliable geometric guidance in early stages, enhance robustness in ill-posed regions, and ultimately improve depth estimation accuracy in the aerial MVS task.

\textbf{Generalization to Different Monocular Priors.}
Apart from recently published DepthAnythingV2~\cite{yang2024depth} and Omnidata~\cite{omnidata}, other monocular depth and normal estimation models can also be integrated into the proposed method to enhance aerial MVS accuracy. To validate this, we conducted experiments replacing DepthAnythingV2 and Omnidata with recent monocular estimation methods available on arXiv, although they have not been officially published yet. Among these, we selected DepthPro~\cite{DepthPro2024} and Lotus~\cite{he2024lotus}, exploring different combinations as monocular priors.
The results, summarized in Table~\ref{tab:gen_mono}, show that ADR-MVS variants trained with different monocular priors (depth + normal) consistently outperform the baseline GoMVS~\cite{wu2024gomvs}. This confirms the effectiveness of our key contribution, the adaptive depth range predictor based on monocular geometric cues, regardless of the specific prior use. Overall, the combination of DepthAnythingV2 and Omnidata remains the optimal choice in terms of both depth estimation accuracy and computational efficiency.
Moreover, while the metric depth estimator DepthPro provides better initialization than the relative depth estimator DepthAnythingV2 (as indicated by lower mean and median deviations in Table~\ref{tab:mono_sta}), models ID-1 and ID-2 achieve higher depth estimation accuracy than models ID-3 and ID-4, respectively. We attribute this to the fact that our depth estimation primarily relies on multi-view feature matching, with the depth range predictor learning monocular depth and normal divergences to construct more discriminative initial cost volume. Consequently, the trained model remains robust with a slightly suboptimal initialization. As illustrated in Fig.~\ref{fig:dif_mono_vis}, the different ADR-MVS models exhibit only minor variations in depth estimation results.

\subsection{Visualization of Point Cloud Reconstructions}
We present the reconstructed point clouds from the WHU and München datasets~\cite{REDNet} \cite{haala2013munchen} in Fig.~\ref{fig:pc_vis}. Overall, the reconstructions exhibit completeness and rich detail across various terrains, demonstrating the advancements of our ADR-MVS in the field of aerial reconstruction.

\section{Conclusion}
In this study, we presented a novel MVS method, ADR-MVS, designed to address the challenges of varying depth ranges and intractable feature matching in aerial imagery. We introduced a depth range predictor that enables adaptive depth range inference using depth and normal estimates, thereby improving the construction of the cost volume at each stage. In the first stage, predicted range maps, derived by exploring the divergence of monocular cues, expand beyond predefined boundaries, enhancing matching discriminability and improving depth estimation accuracy. Over multiple stages, the inferred range maps are progressively narrowed,  leading to more precise depth regression.
Furthermore, we developed a normal-guided cost aggregation technique tailored to the low-detail nature of aerial stereo images, along with a normal-guided refinement module to enhance the final depth map. These contributions effectively utilize monocular geometric cues, improving depth estimation accuracy without incurring additional computational costs. Extensive experiments on the WHU, LuoJia-MVS, and München benchmarks verify the superior performance of ADR-MVS in terms of both accuracy and efficiency.

\bibliographystyle{IEEEtran}
\bibliography{main}

\end{document}